\documentclass[journal]{IEEEtran}

\usepackage{cite}

\usepackage{amssymb,amsfonts}

\usepackage{lipsum}
\usepackage{mathtools}
\usepackage{cuted}

\usepackage{balance}

\usepackage{amsmath}

\DeclareMathOperator*{\esssup}{ess\,sup}

\newtheorem{theorem}{\it Theorem}
\newtheorem{lemma}{\it Lemma}
\newtheorem{definition}{\it Definition}

\newtheorem{corollary}{\it Corollary}
\newtheorem{proposition}{\it Proposition}

\newtheorem{observation}{\it Observation}

\ifCLASSINFOpdf
\else
\fi
\begin{document}
%
\title{Fundamental Limits of Prediction, Generalization, and Recursion: An Entropic-Innovations Perspective
}
%
%
%

\author{Song~Fang~and~Quanyan~Zhu
	\thanks{Song Fang and Quanyan Zhu are with the Department
		of Electrical and Computer Engineering, New York University, USA (e-mail: song.fang@nyu.edu; quanyan.zhu@nyu.edu).}%
}
\maketitle

\begin{abstract}
	In this paper, we examine the fundamental performance limits of prediction, with or without side information. 	
    More specifically, we derive generic lower bounds on the $\mathcal{L}_p$ norms of the prediction errors that are valid for any prediction algorithms and for any data distributions.
    Meanwhile, we combine the entropic analysis from information theory and the innovations approach from prediction/estimation theory to characterize the conditions (in terms of, e.g., directed information or mutual information) to achieve the bounds.
    We also investigate the implications of the results in analyzing the fundamental limits of generalization in fitting (learning) problems from the perspective of prediction with side information, as well as the fundamental limits of recursive algorithms by viewing them as generalized prediction problems.
\end{abstract}


%
\IEEEpeerreviewmaketitle

\section{Introduction}
\label{sec:intro}


\IEEEPARstart{I}{nformation} theory was originally developed to analyze the fundamental limits of communication \cite{shannon1998mathematical, Cov:06}. It turned out that it may also more generally represent any processes that involve ``information transmission'', in its broadest sense, from one point to another; accordingly, the information-theoretic approach has been employed in recent years to analyze the fundamental limits of many systems or problems beyond communication (see, e.g., \cite{mackay2003information,  principe2010information, tishby2015deep, shwartz2017opening, fang2017towards, Kay20, Calin20, rodrigues2021information} and the references therein). Particularly, prediction algorithms (as in prediction theory; see, e.g., \cite{kailath1974view, makhoul1975linear, pourahmadi2001foundations, cesa2006prediction, vaidyanathan2007theory, lindquist2015linear, caines2018linear}) may also be viewed as sequential information transmission processes, broadly speaking, as if extracting as much ``information'' as possible out of all the data points available when making the prediction at each time step, and then transmitting the information to the predicted value, so as to reduce as much as possible the uncertainty or randomness contained in the latter. Note that in the absence of side information, the available data points when making the prediction are merely the past observations, whereas in the presence of side information the available data points include also the (causal) side information in addition to the past observations.
By virtue of this analogy of information transmission, in this paper we examine the fundamental limits of prediction, both with and without side information, via an information-theoretic approach.


In prediction theory, the Kolmogorov--Szeg\"o formula \cite{Pap:02, vaidyanathan2007theory, lindquist2015linear} provides a fundamental bound on the variance of prediction error for the linear prediction (without side information) of Gaussian stochastic processes. In this paper, we go beyond the linear Gaussian case by obtaining generic $\mathcal{L}_{p}$ bounds on the prediction errors, for both the cases with and without side information, that are valid for any prediction algorithms, while the data points can be with arbitrary distributions. 
This is enabled by investigating the underlying entropic relationships of the data points in a sequential manner, allowing the prediction algorithms to be any deterministic or randomized functions/mappings as long as they are causal, while without imposing specific restrictions on the data distributions (e.g., the data points are not necessarily i.i.d.).

In addition, the derived bounds can be characterized explicitly by the conditional entropy of the data point to be predicted given the all the data points, including those from the past observations as well as the (causal) side information, that are available when the prediction is being made.
We also examine the conditions for achieving the lower bounds from an entropic-innovations perspective, that is, in terms of the mutual information between the current innovation \cite{linearestimation} and the previous innovations, as well as the transfer entropy \cite{schreiber2000measuring} and/or directed information \cite{massey1990causality, kramer1998directed, jiao2013universal} from the side information to the innovations process. Accordingly, it is seen in general that one necessary condition to achieve the bounds is that all the usable information from the innovations process as well as the side information has been extracted.
Meanwhile, it is seen that in order to minimize the $\mathcal{L}_p$ norms of the prediction error for different values of $p$, the distributions of the prediction error should be steered to different ones, for instance, Laplace for $p=1$, Gaussian for $p=2$, and uniform for $p = \infty$.
In addition, our lower bounds are seen to reduce to the Kolmogorov--Szeg\"o formula when predicting Gaussian stochastic processes without side information while $p = 2$; on the other hand, they are consistent with the so-called estimation counterparts of Fano's inequality \cite{Cov:06} in the special cases of static estimation problems with and without side information with $p = 2$. As such, the bounds obtained in this paper may be viewed as generalizations of the aforementioned results.


We also analyze the fundamental limits of generalization \cite{hastie2009elements, goodfellow2016deep, wolpert2018mathematics} in learning problems as well as the fundamental limits of recursive algorithms under the general framework of prediction; in the former case, the generalization problem can be treated as a prediction problem with side information, while in the latter, recursive algorithms may be viewed as generalized prediction processes. More specifically, we obtain ``best-case" bounds on the generalization error that are valid for any learning algorithms, in the scenarios of supervised learning \cite{hastie2009elements}, semi-supervised learning \cite{semisupervised}, as well as unsupervised learning \cite{hinton1999unsupervised}. Meanwhile, we derive generic lower bounds on the recursive differences for recursive algorithms \cite{kushner2003stochastic} with arbitrary recursion functions.




The remainder of the paper is organized as follows. Section~II introduces the technical preliminaries. In Section~III, we present the fundamental limits of prediction with and without side information. Section~IV is devoted to the fundamental limits of generalization. In Section~V, we focus on the fundamental limits of recursive algorithms. Concluding remarks are given in Section~VI.

Note that this paper is based mainly upon the conference paper \cite{FangCISS20}, which only discusses prediction without side information. Note as well that prior to \cite{FangCISS20}, the special cases of $p = 2$ and $p = \infty$, both without side information, have been presented respectively in \cite{FangITW19} and \cite{FangMLSP19}; see the discussions in \cite{FangCISS20} for further details.

It is also worth mentioning that an arXiv version of this paper can be found in \cite{FangTSP20arxiv}. Note in particular that it had a different title ``Fundamental Limits of Online Learning: An Entropic-Innovations Viewpoint" for the previous version, which was in fact the same paper, whereas the title was changed to the current one since we think it is more appropriate. This is pointed out herein so as to avoid possible unnecessary confusions to the readers.


\section{Preliminaries}

In this paper, we consider real-valued continuous random variables and vectors, as well as discrete-time stochastic processes they compose. All random variables, random vectors, and stochastic processes are assumed to be zero-mean for simplicity and without loss of generality. We represent random variables and vectors using boldface letters. Given a stochastic process $\left\{ \mathbf{x}_{k}\right\}$, we denote the sequence $\mathbf{x}_0,\ldots,\mathbf{x}_{k}$ by the random vector $\mathbf{x}_{0,\ldots,k}=\left[\mathbf{x}_0^T~\cdots~\mathbf{x}_{k}^T\right]^T$. The logarithm is defined with base $2$. All functions are assumed to be measurable. 
A stochastic process $\left\{ \mathbf{x}_{k}\right\}$ is said to be asymptotically stationary if it is stationary as $k \to \infty$, and herein stationarity means strict stationarity unless otherwise specified \cite{Pap:02}. In addition, a process being asymptotically stationary implies that it is asymptotically mean stationary \cite{gray2011entropy}. 

Note in particular that, for simplicity and
with abuse of notations, we utilize $\mathbf{x} \in \mathbb{R}$ and $\mathbf{x} \in \mathbb{R}^n$ to
indicate that $\mathbf{x}$ is a real-valued random variable and that $\mathbf{x}$
is a real-valued $n$-dimensional random vector, respectively.

Entropy, conditional entropy, and mutual information are the most basic notions in information theory \cite{Cov:06}, which we introduce below.

\begin{definition} The differential entropy of random vector $\mathbf{x}$ with density $p_{\mathbf{x}} \left(x\right)$ is defined as
	\begin{flalign}
	h\left( \mathbf{x} \right)
	=-\int p_{\mathbf{x}} \left(x\right) \log p_{\mathbf{x}} \left(x\right) \mathrm{d} x. \nonumber
	\end{flalign}
	The conditional differential entropy of random vector $\mathbf{x}$ given random vector $\mathbf{y}$ with joint density $p_{\mathbf{x}, \mathbf{y}} \left(x,y\right)$ and conditional density $p_{\mathbf{x} | \mathbf{y}} \left(x,y\right)$ is defined as
	\begin{flalign}
	h\left(\mathbf{x}\middle|\mathbf{y}\right)
	=-\int p_{\mathbf{x}, \mathbf{y}} \left(x,y\right)\log p_{\mathbf{x} | \mathbf{y}} \left(x,y\right) \mathrm{d}x\mathrm{d}y. \nonumber
	\end{flalign}
	The mutual information between random vectors $\mathbf{x}, \mathbf{y}$ with densities $p_{\mathbf{x}} \left(x\right)$, $p_{\mathbf{y}} \left( y \right) $ and joint density $p_{\mathbf{x}, \mathbf{y}} \left(x,y\right)$ is defined as
	\begin{flalign}
	I\left(\mathbf{x};\mathbf{y}\right)
	=\int p_{\mathbf{x}, \mathbf{y}} \left(x,y\right) \log \frac{p_{\mathbf{x}, \mathbf{y}} \left(x,y\right)}{p_{\mathbf{x}} \left(x\right) p_{\mathbf{y}} \left( y \right) }\mathrm{d}x\mathrm{d}y. \nonumber
	\end{flalign}
	The entropy rate of a stochastic process $\left\{ \mathbf{x}_{k}\right\}$ is defined as
	\begin{flalign}
	h_\infty \left(\mathbf{x}\right)=\limsup_{k\to \infty} \frac{h\left(\mathbf{x}_{0,\ldots,k}\right)}{k+1}. \nonumber
	\end{flalign}
\end{definition}

\vspace{0mm}

Properties of these notions can be found in, e.g., \cite{Cov:06}. In particular, the next lemma \cite{dolinar1991maximum} presents the maximum-entropy probability distributions under $\mathcal{L}_{p}$-norm constraints for random variables.

\begin{lemma} \label{maximum}
	Consider a random variable $\mathbf{x} \in \mathbb{R}$ with $\mathcal{L}_{p}$ norm $\left[ \mathbb{E} \left( \left| \mathbf{x} \right|^{p} \right) \right]^{\frac{1}{p}} = \mu,~p \geq 1$.
	Then,  
	\begin{flalign} 
	h \left( \mathbf{x} \right) 
	\leq \log \left[ 2 \Gamma \left( \frac{p+1}{p} \right) \left( p \mathrm{e} \right)^{\frac{1}{p}} \mu \right], \nonumber
	\end{flalign}
	where equality holds if and only if $\mathbf{x}$ is with probability density
	\begin{flalign}
	f_{\mathbf{x}} \left( x \right)
	= \frac{ \mathrm{e}^{- \left| x \right|^{p} / \left( p \mu^{p} \right)} }{2 \Gamma \left( \frac{p+1}{p} \right) p^{\frac{1}{p}} \mu}. \nonumber
	\end{flalign}
	Herein, $\Gamma \left( \cdot \right)$ denotes the Gamma function.
\end{lemma}

In particular, when $p \to \infty$, 
\begin{flalign}
\lim_{p \to \infty} \left[ \mathbb{E} \left( \left| \mathbf{x} \right|^{p} \right) \right]^{\frac{1}{p}} = \esssup_{ f_{\mathbf{x}} \left( x \right) > 0} \left| \mathbf{x} \right|, \nonumber
\end{flalign}
and
\begin{flalign} 
\lim_{p \to \infty} \log \left[ 2 \Gamma \left( \frac{p+1}{p} \right) \left( p \mathrm{e} \right)^{\frac{1}{p}} \mu \right] = \log \left( 2 \mu \right), \nonumber
\end{flalign}
while
\begin{flalign}
\lim_{p \to \infty}\frac{ \mathrm{e}^{- \left| x \right|^{p} / \left( p \mu^{p} \right)} }{2 \Gamma \left( \frac{p+1}{p} \right) p^{\frac{1}{p}} \mu}
= 
\left\{ \begin{array}{cc}
\frac{1}{2 \mu}, & \left| x \right| \leq \mu,\\
0, & \left| x \right| > \mu.
\end{array} \right. \nonumber
\end{flalign}

Meanwhile, the following notions of transfer entropy, directed information, and causally conditional entropy take the issue of causality into consideration \cite{massey1990causality, kramer1998directed, schreiber2000measuring, jiao2013universal}.

\begin{definition}
The transfer entropy from sequence $\left\{ \mathbf{x}_{0, \ldots, k} \right\}$ to sequence $\left\{ \mathbf{y}_{0, \ldots, k} \right\}$ is defined as 
\begin{flalign}
T \left(\mathbf{x}_{0, \ldots, k}  \to \mathbf{y}_{0, \ldots, k} \right)
= I \left( \mathbf{x}_{0,\ldots,k} ; \mathbf{y}_{k} | \mathbf{y}_{0,\ldots,k-1} \right).\nonumber
\end{flalign}
The directed information from sequence $\left\{ \mathbf{x}_{0, \ldots, k} \right\}$ to sequence $\left\{ \mathbf{y}_{0, \ldots, k} \right\}$ is defined as 
\begin{flalign}
I \left(\mathbf{x}_{0, \ldots, k}  \to \mathbf{y}_{0, \ldots, k} \right)
= \sum_{i=0}^{k} I \left( \mathbf{x}_{0,\ldots,i} ; \mathbf{y}_{i} | \mathbf{y}_{0,\ldots,i-1} \right).\nonumber
\end{flalign}
The causally conditional entropy of sequence $\left\{ \mathbf{y}_{0, \ldots, k} \right\}$ given sequence $\left\{ \mathbf{x}_{0, \ldots, k} \right\}$ is defined as
\begin{flalign}
h \left(\mathbf{y}_{0, \ldots, k} \| \mathbf{x}_{0, \ldots, k} \right)
&= \sum_{i=0}^{k} h \left(\mathbf{y}_{i} | \mathbf{y}_{0,\ldots,i-1}, \mathbf{x}_{0,\ldots,i}\right). \nonumber
\end{flalign}
The transfer entropy rate from stochastic process $\left\{ \mathbf{x}_{k} \right\}$ to process $\left\{ \mathbf{y}_{k} \right\}$ is defined as 
\begin{flalign}
T_{\infty} \left(\mathbf{x} \to \mathbf{y}\right)
= \limsup_{k\to \infty} I \left( \mathbf{x}_{0,\ldots,k} ; \mathbf{y}_{k} | \mathbf{y}_{0,\ldots,k-1} \right). \nonumber
\end{flalign}
The directed information rate from stochastic process $\left\{ \mathbf{x}_{k} \right\}$ to process $\left\{ \mathbf{y}_{k} \right\}$ is defined as
\begin{flalign}
I_{\infty}\left(\mathbf{x} \to \mathbf{y}\right)
&= \limsup_{k\to \infty} \frac{\sum_{i=0}^{k} I \left( \mathbf{x}_{0,\ldots,i} ; \mathbf{y}_{i} | \mathbf{y}_{0,\ldots,i-1} \right)}{k+1}. \nonumber
\end{flalign}
The causally conditional information rate of stochastic process $\left\{ \mathbf{y}_{k} \right\}$ given process $\left\{ \mathbf{x}_{k} \right\}$ is defined as
\begin{flalign}
h_{\infty}\left(\mathbf{y} \| \mathbf{x}\right)
&= \limsup_{k\to \infty} \frac{\sum_{i=0}^{k} h \left(\mathbf{y}_{i} | \mathbf{y}_{0,\ldots,i-1}, \mathbf{x}_{0,\ldots,i}\right)}{k+1}. \nonumber
\end{flalign}
\end{definition}

\section{Fundamental Limits of Prediction with or without Side Information} \label{estimation}

In this section, we investigate the fundamental limits of prediction, for both the scenarios with and without side information. We consider the case with side information first.

Specifically, consider the input/output pairs $\left( \mathbf{x}_{i}, \mathbf{y}_{i} \right)$, $i = 0, \ldots, k$, where $\mathbf{x}_{i} \in \mathbb{R}^n$ denotes the input while $\mathbf{y}_{i} \in \mathbb{R}$ denotes the output. Suppose that at time $k$, the trained prediction algorithm (based upon all the previous input/output pairs $\left( \mathbf{x}_{i}, \mathbf{y}_{i} \right)$, $i = 0, \ldots, k - 1$), as a mapping from input to output, is denoted as $g_{k} \left( \cdot \right)$. (Note in particular that  the true values of $ \mathbf{x}_{0,\ldots,k-1}$, $\mathbf{y}_{0,\ldots,k-1}$, and $\mathbf{x}_{k}$ are known at time $k$, whereas the true value of $\mathbf{y}_{k}$ is not known until time $k+1$.) Then, $g_{k} \left( \cdot \right)$ will be employed to give a prediction of $\mathbf{y}_{k}$ with input $\mathbf{x}_{k}$, and this prediction is denoted as 
\begin{flalign}
\widehat{\mathbf{y}}_{k} = g_{k} \left( \mathbf{x}_{k} \right).
\end{flalign}

In what follows, we  derive generic bounds on the $\mathcal{L}_{p}$ norm of the prediction error $\mathbf{y}_{k} - \widehat{\mathbf{y}}_{k}$, which are valid for all possible prediction algorithms $g_{k} \left( \cdot \right)$, as any deterministic or randomized functions/mappings. Towards this end, we shall first present the following observation.

\begin{observation} \label{o1}
	Note that the parameters of $g_{k} \left( \cdot \right)$ are trained using $ \left( \mathbf{x}_{i}, \mathbf{y}_{i} \right), i = 0,\ldots,k-1$, hence eventually it holds that
	\begin{flalign} \label{predction}
	\widehat{\mathbf{y}}_{k} 
	&= g_{k} \left( \mathbf{x}_{k} \right) = \widehat{g}_{k} \left( \mathbf{x}_{k}, \mathbf{y}_{0,\ldots,k-1}, \mathbf{x}_{0,\ldots,k-1} \right) \nonumber \\
	& = \widehat{g}_{k} \left( \mathbf{y}_{0,\ldots,k-1}, \mathbf{x}_{0,\ldots,k} \right),
	\end{flalign}
	indicating that $\widehat{\mathbf{y}}_{k}$ is ultimately a function, denoted herein as $\widehat{g}_{k} \left( \cdot \right)$, of $ \mathbf{x}_{k}, \mathbf{y}_{0,\ldots,k-1}, \mathbf{x}_{0,\ldots,k-1}$ (or equivalently, $\mathbf{y}_{0,\ldots,k-1}, \mathbf{x}_{0,\ldots,k}$), representing all the data that is available at time $k$. 
\end{observation}

As such, \eqref{predction}  essentially features a prediction problem with causal side information, that is, to predict $\mathbf{y}_{k}$ based on the past $\mathbf{y}_{0,\ldots,k-1}$ with causal side information $\mathbf{x}_{0,\ldots,k}$. This is a key observation that enables obtaining the subsequent result.

\begin{theorem} \label{MIMOFano}
	For any prediction algorithm $g_{k} \left( \cdot \right)$, it holds that  
	\begin{flalign} \label{MIMOestimation}
	\left[ \mathbb{E} \left( \left| \mathbf{y}_{k} - \widehat{\mathbf{y}}_{k} \right|^{p} \right) \right]^{\frac{1}{p}}
	\geq \frac{2^{h \left( \mathbf{y}_k | \mathbf{y}_{0,\ldots,k-1}, \mathbf{x}_{0,\ldots,k} \right)}}{2 \Gamma \left( \frac{p+1}{p} \right) \left( p \mathrm{e} \right)^{\frac{1}{p}}},
	\end{flalign}
	where equality holds if and only if $\mathbf{y}_{k} - \widehat{\mathbf{y}}_{k}$ is with probability density
	\begin{flalign} \label{estimationdistribution}
	f_{\mathbf{y}_{k} - \widehat{\mathbf{y}}_{k}} \left( y \right)
	= \frac{ \mathrm{e}^{- \left| y \right|^{p} / \left( p \mu^{p} \right)} }{2 \Gamma \left( \frac{p+1}{p} \right) p^{\frac{1}{p}} \mu},
	\end{flalign}
	and $I \left( \mathbf{y}_{k} - \widehat{\mathbf{y}}_{k}; \mathbf{y}_{0,\ldots,k-1}, \mathbf{x}_{0,\ldots,k} \right) = 0$. Note that herein
	\begin{flalign}
	\mu 
	= \frac{2^{h \left( \mathbf{y}_k | \mathbf{y}_{0,\ldots,k-1}, \mathbf{x}_{0,\ldots,k} \right)}}{2 \Gamma \left( \frac{p+1}{p} \right) \left( p \mathrm{e} \right)^{\frac{1}{p}}}.
	\end{flalign}
\end{theorem}

\begin{IEEEproof}
	It is known from Lemma~\ref{maximum} that 
	\begin{flalign} 
	\left[ \mathbb{E} \left( \left| \mathbf{y}_{k} - \widehat{\mathbf{y}}_{k} \right|^{p} \right) \right]^{\frac{1}{p}}
	\geq \frac{2^{h \left(  \mathbf{y}_{k} - \widehat{\mathbf{y}}_{k} \right)}}{2 \Gamma \left( \frac{p+1}{p} \right) \left( p \mathrm{e} \right)^{\frac{1}{p}}}, \nonumber
	\end{flalign} 
	where equality holds if and only if $\mathbf{y}_{k} - \widehat{\mathbf{y}}_{k}$ is with probability density \eqref{estimationdistribution}. Meanwhile,
	\begin{flalign} 
	&h \left( \mathbf{y}_{k} - \widehat{\mathbf{y}}_{k} \right)  \nonumber \\
	&~~~~ = h \left( \mathbf{y}_{k} - \widehat{\mathbf{y}}_{k} |  \mathbf{y}_{0,\ldots,k-1}, \mathbf{x}_{0,\ldots,k} \right) \nonumber \\
	&~~~~~~~~ + I \left( \mathbf{y}_{k} - \widehat{\mathbf{y}}_{k}; \mathbf{y}_{0,\ldots,k-1}, \mathbf{x}_{0,\ldots,k} \right) \nonumber \\
	&~~~~= h \left( \mathbf{y}_k - g_{k} \left(  \mathbf{y}_{0,\ldots,k-1}, \mathbf{x}_{0,\ldots,k} \right) |  \mathbf{y}_{0,\ldots,k-1}, \mathbf{x}_{0,\ldots,k} \right) \nonumber \\
	&~~~~~~~~  + I \left( \mathbf{y}_{k} - \widehat{\mathbf{y}}_{k}; \mathbf{y}_{0,\ldots,k-1}, \mathbf{x}_{0,\ldots,k} \right) \nonumber \\
	&~~~~= h \left( \mathbf{y}_k | \mathbf{y}_{0,\ldots,k-1}, \mathbf{x}_{0,\ldots,k} \right) \nonumber \\
	&~~~~~~~~ + I \left( \mathbf{y}_{k} - \widehat{\mathbf{y}}_{k}; \mathbf{y}_{0,\ldots,k-1}, \mathbf{x}_{0,\ldots,k} \right) \nonumber \\
	&~~~~ \geq h \left( \mathbf{y}_k | \mathbf{y}_{0,\ldots,k-1}, \mathbf{x}_{0,\ldots,k} \right). \nonumber
	\end{flalign}
	As a result,
	\begin{flalign}
	2^{ h \left( \mathbf{y}_{k} - \widehat{\mathbf{y}}_{k} \right)} 
	\geq 2^{h \left( \mathbf{y}_k | \mathbf{y}_{0,\ldots,k-1}, \mathbf{x}_{0,\ldots,k} \right)}, \nonumber
	\end{flalign}
	where equality holds if and only if $I \left( \mathbf{y}_{k} - \widehat{\mathbf{y}}_{k}; \mathbf{y}_{0,\ldots,k-1}, \mathbf{x}_{0,\ldots,k} \right) = 0$.
	Therefore, \eqref{MIMOestimation} follows, where equality holds if and only if $\mathbf{y}_{k} - \widehat{\mathbf{y}}_{k}$ is with probability density \eqref{estimationdistribution} and $I \left( \mathbf{y}_{k} - \widehat{\mathbf{y}}_{k}; \mathbf{y}_{0,\ldots,k-1}, \mathbf{x}_{0,\ldots,k} \right) = 0$.
\end{IEEEproof}


In general, it is seen that, for all $p \geq 1$, the lower bound in \eqref{MIMOestimation} depends only on the conditional entropy (the amount of ``randomness'') of the output $\mathbf{y}_{k}$ to be predicted given the corresponding input $\mathbf{x}_{k}$ as well as all the previous inputs $\mathbf{x}_{0,\ldots,k-1}$ and outputs $\mathbf{y}_{0,\ldots,k-1}$. Accordingly, if $\mathbf{y}_{0,\ldots,k-1}, \mathbf{x}_{0,\ldots,k}$ provide more/less information of $\mathbf{y}_{k}$, then the lower bound becomes smaller/larger. 

On the other hand, equality in \eqref{MIMOestimation} holds if and only if the prediction error $\mathbf{y}_{k} - \widehat{\mathbf{y}}_{k}$ is with probability \eqref{estimationdistribution}, and contains no information of $\mathbf{y}_{0,\ldots,k-1}, \mathbf{x}_{0,\ldots,k}$; it is as if all the ``information'' (from all the data available when making the prediction) that may be utilized to reduce the $\mathcal{L}_{p}$ norm of the prediction error has been extracted. This is more clearly seen from the viewpoint of ``entropic innovations'', as will be discussed subsequently.



\begin{proposition} \label{p2}
	For any $g_{k} \left( \cdot \right)$, it holds that  
	\begin{flalign} \label{innovations}
	&I \left( \mathbf{y}_{k} - \widehat{\mathbf{y}}_{k}; \mathbf{y}_{0,\ldots,k-1}, \mathbf{x}_{0,\ldots,k} \right) \nonumber \\
	&~~~~ = I \left( \mathbf{y}_{k} - \widehat{\mathbf{y}}_{k} ; \mathbf{y}_{0} - \widehat{\mathbf{y}}_{0}, \ldots, \mathbf{y}_{k-1} - \widehat{\mathbf{y}}_{k-1}, \mathbf{x}_{0,\ldots,k} \right)
	\nonumber \\
	&~~~~ = I \left( \mathbf{y}_{k} - \widehat{\mathbf{y}}_{k} ; \mathbf{y}_{0} - \widehat{\mathbf{y}}_{0}, \ldots, \mathbf{y}_{k-1} - \widehat{\mathbf{y}}_{k-1} \right) \nonumber \\
	&~~~~~~~~ + I \left( \mathbf{x}_{0,\ldots,k} ; \mathbf{y}_{k} - \widehat{\mathbf{y}}_{k} | \mathbf{y}_{0} - \widehat{\mathbf{y}}_{0}, \ldots, \mathbf{y}_{k-1} - \widehat{\mathbf{y}}_{k-1} \right).
	\end{flalign}
\end{proposition}

\begin{IEEEproof}
	Since $\widehat{\mathbf{y}}_{k-1} = \widehat{g}_{k-1} \left( \mathbf{y}_{0,\ldots,k-2}, \mathbf{x}_{0,\ldots,k-1} \right) $, we have (by the data processing inequality \cite{Cov:06})
	\begin{flalign} 
	&I \left( \mathbf{y}_{k} - \widehat{\mathbf{y}}_{k} ; \mathbf{y}_{0,\ldots,k-1}, \mathbf{x}_{0,\ldots,k} \right) \nonumber \\
	&~~~~ = I \left( \mathbf{y}_{k} - \widehat{\mathbf{y}}_{k} ; \mathbf{y}_{0,\ldots,k-2}, \mathbf{y}_{k-1} - \widehat{\mathbf{y}}_{k-1}, \mathbf{x}_{0,\ldots,k} \right).\nonumber
	\end{flalign}
	As such, by invoking the data processing inequality repeatedly, it follows that
	\begin{flalign} 
	&I \left( \mathbf{y}_{k} - \widehat{\mathbf{y}}_{k} ; \mathbf{y}_{0,\ldots,k-2}, \mathbf{y}_{k-1} - \widehat{\mathbf{y}}_{k-1}, \mathbf{x}_{0,\ldots,k} \right) \nonumber \\
	& = I \left( \mathbf{y}_{k} - \widehat{\mathbf{y}}_{k} ; \mathbf{y}_{0,\ldots,k-3}, \mathbf{y}_{k-2} - \widehat{\mathbf{y}}_{k-2}, \mathbf{y}_{k-1} - \widehat{\mathbf{y}}_{k-1}, \mathbf{x}_{0,\ldots,k} \right) 
	\nonumber \\
	& = \cdots = I \left( \mathbf{y}_{k} - \widehat{\mathbf{y}}_{k} ; \mathbf{y}_{0} - \widehat{\mathbf{y}}_{0}, \ldots, \mathbf{y}_{k-1} - \widehat{\mathbf{y}}_{k-1}, \mathbf{x}_{0,\ldots,k} \right). \nonumber
	\end{flalign}
	Eventually, this leads to
	\begin{flalign} 
	&I \left( \mathbf{y}_{k} - \widehat{\mathbf{y}}_{k} ; \mathbf{y}_{0,\ldots,k-1}, \mathbf{x}_{0,\ldots,k} \right) \nonumber \\
	&~~~~ =I \left( \mathbf{y}_{k} - \widehat{\mathbf{y}}_{k} ; \mathbf{y}_{0} - \widehat{\mathbf{y}}_{0}, \ldots, \mathbf{y}_{k-1} - \widehat{\mathbf{y}}_{k-1}, \mathbf{x}_{0,\ldots,k} \right). \nonumber
	\end{flalign}
	On the other hand,
	\begin{flalign} 
	&I \left( \mathbf{y}_{k} - \widehat{\mathbf{y}}_{k} ; \mathbf{y}_{0} - \widehat{\mathbf{y}}_{0}, \ldots, \mathbf{y}_{k-1} - \widehat{\mathbf{y}}_{k-1}, \mathbf{x}_{0,\ldots,k} \right)\nonumber \\
	&~~~~ = I \left( \mathbf{y}_{k} - \widehat{\mathbf{y}}_{k} ; \mathbf{y}_{0} - \widehat{\mathbf{y}}_{0}, \ldots, \mathbf{y}_{k-1} - \widehat{\mathbf{y}}_{k-1} \right) \nonumber \\
	&~~~~~~~~ + I \left( \mathbf{x}_{0,\ldots,k} ; \mathbf{y}_{k} - \widehat{\mathbf{y}}_{k} | \mathbf{y}_{0} - \widehat{\mathbf{y}}_{0}, \ldots, \mathbf{y}_{k-1} - \widehat{\mathbf{y}}_{k-1} \right). \nonumber
	\end{flalign}
	This completes the proof.
\end{IEEEproof}

On the right-hand side of \eqref{innovations}, the first term $I \left( \mathbf{y}_{k} - \widehat{\mathbf{y}}_{k} ; \mathbf{y}_{0} - \widehat{\mathbf{y}}_{0}, \ldots, \mathbf{y}_{k-1} - \widehat{\mathbf{y}}_{k-1} \right)$ denotes the mutual information between the current ``innovation" \cite{linearestimation} $\mathbf{y}_{k} - \widehat{\mathbf{y}}_{k}$ and the past innovations $\mathbf{y}_{0} - \widehat{\mathbf{y}}_{0}, \ldots, \mathbf{y}_{k-1} - \widehat{\mathbf{y}}_{k-1}$. Meanwhile, the second term $ 
I \left( \mathbf{x}_{0,\ldots,k} ; \mathbf{y}_{k} - \widehat{\mathbf{y}}_{k} | \mathbf{y}_{0} - \widehat{\mathbf{y}}_{0}, \ldots, \mathbf{y}_{k-1} - \widehat{\mathbf{y}}_{k-1} \right)
$
represents the transfer entropy \cite{schreiber2000measuring} from the input process $\mathbf{x}_{0,\ldots,k} $ to the innovations process $\left( \mathbf{y}_{0} - \widehat{\mathbf{y}}_{0}, \ldots, \mathbf{y}_{k} - \widehat{\mathbf{y}}_{k} \right) $ from time $0$ to $k$, which is denoted as 
\begin{flalign}
&T \left( \mathbf{x}_{0,\ldots,k} \to \left( \mathbf{y}_{0} - \widehat{\mathbf{y}}_{0}, \ldots, \mathbf{y}_{k} - \widehat{\mathbf{y}}_{k} \right) \right) \nonumber \\
&~~~~ = I \left( \mathbf{x}_{0,\ldots,k} ; \mathbf{y}_{k} - \widehat{\mathbf{y}}_{k} | \mathbf{y}_{0} - \widehat{\mathbf{y}}_{0}, \ldots, \mathbf{y}_{k-1} - \widehat{\mathbf{y}}_{k-1} \right).
\end{flalign}
As such, the necessary and sufficient condition for 
\begin{flalign} \label{zerocondition}
I \left( \mathbf{y}_{k} - \widehat{\mathbf{y}}_{k}; \mathbf{y}_{0,\ldots,k-1}, \mathbf{x}_{0,\ldots,k} \right) = 0
\end{flalign}
is that
\begin{flalign} \label{zerocondition2}
I \left( \mathbf{y}_{k} - \widehat{\mathbf{y}}_{k} ; \mathbf{y}_{0} - \widehat{\mathbf{y}}_{0}, \ldots, \mathbf{y}_{k-1} - \widehat{\mathbf{y}}_{k-1} \right) = 0
\end{flalign}
while simultaneously
\begin{flalign} \label{zerocondtion3}
T \left( \mathbf{x}_{0,\ldots,k} \to \left( \mathbf{y}_{0} - \widehat{\mathbf{y}}_{0}, \ldots, \mathbf{y}_{k} - \widehat{\mathbf{y}}_{k} \right) \right) = 0.
\end{flalign}
Note that \eqref{zerocondition2} implies that the current innovation contains no information of the previous innovations, meaning that the information that can be utilized has been extracted from the innovations process. Meanwhile, \eqref{zerocondtion3} mandates that no information is being (directedly) transferred from the input process to the innovation process \cite{schreiber2000measuring}, implicating that all the usable information from the input process has been squeezed out. As such, \eqref{zerocondition} indicates that all the information from the input and the output processes, which are the two and only two ultimate sources of information, has been made use of.

Proposition~\ref{p2} is a key link that facilitates the subsequent analysis in the asymptotic case.

\begin{corollary} \label{MIMOasymp}
	For any prediction algorithm $g_{k} \left( \cdot \right)$, it holds that  
	\begin{flalign} \label{MIMOasymp1}
	\liminf_{k\to \infty} \left[ \mathbb{E} \left( \left| \mathbf{y}_{k} - \widehat{\mathbf{y}}_{k} \right|^{p} \right) \right]^{\frac{1}{p}}
	\geq \liminf_{k\to \infty} \frac{2^{h \left( \mathbf{y}_k | \mathbf{y}_{0,\ldots,k-1}, \mathbf{x}_{0,\ldots,k} \right)}}{2 \Gamma \left( \frac{p+1}{p} \right) \left( p \mathrm{e} \right)^{\frac{1}{p}}},
	\end{flalign}
	where equality holds if $\left\{ \mathbf{y}_{k} - \widehat{\mathbf{y}}_{k} \right\}$ is asymptotically independent over time, i.e.,
	\begin{flalign}
	\lim_{k \to \infty} I \left( \mathbf{y}_{k} - \widehat{\mathbf{y}}_{k} ; \mathbf{y}_{0} - \widehat{\mathbf{y}}_{0}, \ldots, \mathbf{y}_{k-1} - \widehat{\mathbf{y}}_{k-1} \right) = 0,
	\end{flalign} 
	and with probability density
	\begin{flalign} \label{asydistribution}
	\lim_{k \to \infty} f_{\mathbf{y}_{k} - \widehat{\mathbf{y}}_{k}} \left( y \right)
	= \frac{ \mathrm{e}^{- \left| y \right|^{p} / \left( p \mu^{p} \right)} }{2 \Gamma \left( \frac{p+1}{p} \right) p^{\frac{1}{p}} \mu},
	\end{flalign}
	while the directed information rate from $\left\{ \mathbf{x}_{k} \right\}$ to $\left\{ \mathbf{y}_{k} - \widehat{\mathbf{y}}_{k} \right\}$ is asymptotically zero, i.e., 
	\begin{flalign} 
	T_{\infty} \left( \mathbf{x} \to \left( \mathbf{y} - \widehat{\mathbf{y}} \right) \right) = 0. 
	\end{flalign}
	Note that herein
	\begin{flalign}
	\mu 
	= \liminf_{k\to \infty} \frac{2^{h \left( \mathbf{y}_k | \mathbf{y}_{0,\ldots,k-1}, \mathbf{x}_{0,\ldots,k} \right)}}{2 \Gamma \left( \frac{p+1}{p} \right) \left( p \mathrm{e} \right)^{\frac{1}{p}}}.
	\end{flalign}
\end{corollary}

\begin{IEEEproof}
	As in the proof of Theorem~\ref{MIMOFano}, it can be shown that
	\begin{flalign} 
	\left[ \mathbb{E} \left( \left| \mathbf{y}_{k} - \widehat{\mathbf{y}}_{k} \right|^{p} \right) \right]^{\frac{1}{p}}
	\geq \frac{2^{h \left( \mathbf{y}_k | \mathbf{y}_{0,\ldots,k-1}, \mathbf{x}_{0,\ldots,k} \right)}}{2 \Gamma \left( \frac{p+1}{p} \right) \left( p \mathrm{e} \right)^{\frac{1}{p}}}, \nonumber
	\end{flalign} 
	where equality holds if and only if $\mathbf{y}_{k} - \widehat{\mathbf{y}}_{k}$ is with probability density \eqref{estimationdistribution} and $I \left( \mathbf{y}_{k} - \widehat{\mathbf{y}}_{k}; \mathbf{y}_{0,\ldots,k-1}, \mathbf{x}_{0,\ldots,k} \right) = 0$. This, by taking $\liminf_{k\to \infty}$ on both sides, then leads to
	\begin{flalign} 
	\liminf_{k\to \infty} \left[ \mathbb{E} \left( \left| \mathbf{y}_{k} - \widehat{\mathbf{y}}_{k} \right|^{p} \right) \right]^{\frac{1}{p}}
	\geq \liminf_{k\to \infty} \frac{2^{h \left( \mathbf{y}_k | \mathbf{y}_{0,\ldots,k-1}, \mathbf{x}_{0,\ldots,k} \right)}}{2 \Gamma \left( \frac{p+1}{p} \right) \left( p \mathrm{e} \right)^{\frac{1}{p}}}. \nonumber
	\end{flalign}
	Herein, equality holds if $\left\{ \mathbf{y}_{k} - \widehat{\mathbf{y}}_{k} \right\}$ is asymptotically with probability density \eqref{asydistribution} and 
	\begin{flalign}
	\lim_{k\to \infty} I \left( \mathbf{y}_{k} - \widehat{\mathbf{y}}_{k}; \mathbf{y}_{0,\ldots,k-1}, \mathbf{x}_{0,\ldots,k} \right) = 0, \nonumber
	\end{flalign}
	which in turn holds if (cf. discussions after Theorem~\ref{MIMOFano})
	\begin{flalign} 
	\lim_{k\to \infty} I \left( \mathbf{y}_{k} - \widehat{\mathbf{y}}_{k} ; \mathbf{y}_{0} - \widehat{\mathbf{y}}_{0}, \ldots, \mathbf{y}_{k-1} - \widehat{\mathbf{y}}_{k-1} \right)= 0 \nonumber
	\end{flalign}
	while at the same time
	\begin{flalign} 
	&T_{\infty} \left( \mathbf{x} \to \left( \mathbf{y} - \widehat{\mathbf{y}} \right) \right) \nonumber \\ 
	&~~~~ 
	= \limsup_{k\to \infty} I \left( \mathbf{x}_{0,\ldots,k} ; \mathbf{y}_{k} - \widehat{\mathbf{y}}_{k} | \mathbf{y}_{0} - \widehat{\mathbf{y}}_{0}, \ldots, \mathbf{y}_{k-1} - \widehat{\mathbf{y}}_{k-1} \right) \nonumber \\
	&~~~~ = 0. \nonumber
	\end{flalign}
	In other words, equality in \eqref{MIMOasymp1} holds if $\left\{ \mathbf{y}_{k} - \widehat{\mathbf{y}}_{k} \right\}$ is asymptotically independent over time and with probability density \eqref{asydistribution} while the directed information rate from $\left\{ \mathbf{x}_{k} \right\}$ to $\left\{ \mathbf{y}_{k} - \widehat{\mathbf{y}}_{k} \right\}$ is asymptotically zero.
\end{IEEEproof}


In what follows, we consider the special case of asymptotically stationary processes. We shall first show the following result.

\begin{proposition} When $\left\{ \mathbf{x}_{k} \right\}$ and $\left\{ \mathbf{y}_{k} - \widehat{\mathbf{y}}_{k}  \right\}$ are asymptotically stationary, it holds that
	\begin{flalign} \label{directed}
	T_{\infty} \left( \mathbf{x} \to \left( \mathbf{y} - \widehat{\mathbf{y}} \right) \right) = I_{\infty} \left( \mathbf{x} \to \left( \mathbf{y} - \widehat{\mathbf{y}} \right) \right),
	\end{flalign}
	where $I_{\infty} \left( \mathbf{x} \to \left( \mathbf{y} - \widehat{\mathbf{y}} \right) \right)$ denotes the directed information rate \cite{massey1990causality, kramer1998directed, jiao2013universal} from the input process $\left\{ \mathbf{x}_{k} \right\}$ to the innovations process $\left\{ \mathbf{y}_{k} - \widehat{\mathbf{y}}_{k} \right\}$. 
\end{proposition}

\begin{IEEEproof}
	Note that for asymptotically stationary $\left\{ \mathbf{x}_{k} \right\}$ and $\left\{ \mathbf{y}_{k} - \widehat{\mathbf{y}}_{k}  \right\}$, we have
	\begin{flalign}
	&T_{\infty} \left( \mathbf{x} \to \left( \mathbf{y} - \widehat{\mathbf{y}} \right) \right) \nonumber \\ 
	&~~~~ 
	= \limsup_{k\to \infty} I \left( \mathbf{x}_{0,\ldots,k} ; \mathbf{y}_{k} - \widehat{\mathbf{y}}_{k} | \mathbf{y}_{0} - \widehat{\mathbf{y}}_{0}, \ldots, \mathbf{y}_{k-1} - \widehat{\mathbf{y}}_{k-1} \right) \nonumber \\
	&~~~~ = \lim_{k\to \infty} I \left( \mathbf{x}_{0,\ldots,k} ; \mathbf{y}_{k} - \widehat{\mathbf{y}}_{k} | \mathbf{y}_{0} - \widehat{\mathbf{y}}_{0}, \ldots, \mathbf{y}_{k-1} - \widehat{\mathbf{y}}_{k-1} \right) \nonumber \\
	&~~~~ = \lim_{k\to \infty} \frac{\sum_{i=0}^{k} I \left( \mathbf{x}_{0,\ldots,i} ; \mathbf{y}_{i} - \widehat{\mathbf{y}}_{i} | \widehat{\mathbf{y}}_{0}, \ldots, \mathbf{y}_{i-1} - \widehat{\mathbf{y}}_{i-1} \right)}{k+1} \nonumber \\
	&~~~~ = \limsup_{k\to \infty} \frac{\sum_{i=0}^{k} I \left( \mathbf{x}_{0,\ldots,i} ; \mathbf{y}_{i} - \widehat{\mathbf{y}}_{i} | \widehat{\mathbf{y}}_{0}, \ldots, \mathbf{y}_{i-1} - \widehat{\mathbf{y}}_{i-1} \right)}{k+1} \nonumber \\
	&~~~~ = I_{\infty} \left( \mathbf{x} \to \left( \mathbf{y} - \widehat{\mathbf{y}} \right) \right). \nonumber
	\end{flalign}
	This completes the proof.
\end{IEEEproof}

As a consequence, we may arrive at the following result.

\begin{corollary} \label{uniform}
	Assume that $\left\{ \mathbf{x}_{k} \right\}$ and $\left\{ \mathbf{y}_{k} \right\}$ are asymptotically stationary. Then, it holds for any prediction algorithm $g_{k} \left( \cdot \right)$ that 
	\begin{flalign} 
	\liminf_{k\to \infty} \left[ \mathbb{E} \left( \left| \mathbf{y}_{k} - \widehat{\mathbf{y}}_{k} \right|^{p} \right) \right]^{\frac{1}{p}}
	\geq \frac{2^{h_{\infty} \left( \mathbf{y} \| \mathbf{x} \right)}}{2 \Gamma \left( \frac{p+1}{p} \right) \left( p \mathrm{e} \right)^{\frac{1}{p}}},
	\end{flalign}
	where 
	\begin{flalign} 
	h_{\infty} \left( \mathbf{y} \| \mathbf{x} \right) = \lim_{k\to \infty} h \left( \mathbf{y}_k | \mathbf{y}_{0,\ldots,k-1}, \mathbf{x}_{0,\ldots,k} \right)
	\end{flalign}
	denotes the causally conditional entropy rate \cite{kramer1998directed} of $\left\{ \mathbf{y}_{k} \right\}$ given $\left\{ \mathbf{x}_{k} \right\}$. Herein, equality holds if $\left\{ \mathbf{y}_{k} - \widehat{\mathbf{y}}_{k} \right\}$ 
	asymptotically independent over time and with probability density \eqref{asydistribution}
	while the directed information rate from $\left\{ \mathbf{x}_{k} \right\}$ to  $\left\{ \mathbf{y}_{k} - \widehat{\mathbf{y}}_{k} \right\}$ is zero, i.e.,
	\begin{flalign} \label{informationflow}
	I_{\infty} \left( \mathbf{x} \to \left( \mathbf{y} - \widehat{\mathbf{y}} \right) \right) = 0.
	\end{flalign}
\end{corollary}

\begin{IEEEproof}
	Corollary~\ref{uniform} follows directly from Corollary~\ref{MIMOasymp} by noting that for asymptotically stationary processes $\left\{ \mathbf{x}_{k} \right\}$ and $\left\{ \mathbf{y}_{k} \right\}$, we have \cite{Cov:06} 
	\begin{flalign} 
	&\liminf_{k\to \infty} h \left( \mathbf{y}_k | \mathbf{y}_{0,\ldots,k-1}, \mathbf{x}_{0,\ldots,k} \right) \nonumber \\
	&~~~~ = \lim_{k\to \infty} h \left( \mathbf{y}_k | \mathbf{y}_{0,\ldots,k-1}, \mathbf{x}_{0,\ldots,k} \right) \nonumber \\
	&~~~~
	= \lim_{k\to \infty} \frac{\sum_{i=0}^{k} h \left(\mathbf{y}_{i} | \mathbf{y}_{0,\ldots,i-1}, \mathbf{x}_{0,\ldots,i}\right)}{k+1}
	\nonumber \\
	&~~~~
	= \limsup_{k\to \infty} \frac{\sum_{i=0}^{k} h \left(\mathbf{y}_{i} | \mathbf{y}_{0,\ldots,i-1}, \mathbf{x}_{0,\ldots,i}\right)}{k+1}
	= h_{\infty} \left( \mathbf{y} \| \mathbf{x} \right). \nonumber
	\end{flalign}
	Meanwhile, note that if $\left\{ \mathbf{y}_{k} - \widehat{\mathbf{y}}_{k} \right\}$ is
	asymptotically with probability density \eqref{asydistribution}, indicating that it is
	asymptotically stationary, then \eqref{directed} holds since $\left\{ \mathbf{x}_{k} \right\}$ is also asymptotically stationary. This completes the proof.
\end{IEEEproof}

Note that herein \eqref{informationflow} means that the information flow \cite{massey1990causality, kramer1998directed, jiao2013universal} from $\left\{ \mathbf{x}_{k} \right\}$ to  $\left\{ \mathbf{y}_{k} - \widehat{\mathbf{y}}_{k} \right\}$ is zero.

In addition, if $\left\{ \mathbf{y}_{k} - \widehat{\mathbf{y}}_{k} \right\}$ is
asymptotically independent over time and with probability density \eqref{asydistribution} while  the directed information rate from $\left\{ \mathbf{x}_{k} \right\}$ to  $\left\{ \mathbf{y}_{k} - \widehat{\mathbf{y}}_{k} \right\}$ is zero, then, noting also that $\left\{ \mathbf{x}_{k} \right\}$ and $\left\{ \mathbf{y}_{k} \right\}$ are asymptotically stationary, it holds that
\begin{flalign} \label{equality}
\lim_{k\to \infty} \left[ \mathbb{E} \left( \left| \mathbf{y}_{k} - \widehat{\mathbf{y}}_{k} \right|^{p} \right) \right]^{\frac{1}{p}}
= \frac{2^{h_{\infty} \left( \mathbf{y} \| \mathbf{x} \right)}}{2 \Gamma \left( \frac{p+1}{p} \right) \left( p \mathrm{e} \right)^{\frac{1}{p}}}.
\end{flalign}
In fact, we can show that \eqref{equality} holds if and only if  $\left\{ \mathbf{y}_{k} - \widehat{\mathbf{y}}_{k} \right\}$ is asymptotically independent over time and with probability density \eqref{asydistribution} while the directed information rate from $\left\{ \mathbf{x}_{k} \right\}$ to  $\left\{ \mathbf{y}_{k} - \widehat{\mathbf{y}}_{k} \right\}$ is zero.

\subsection{Prediction without Side Information}

We now consider the case without side information. Particularly, when the causal side information $\mathbf{x}_{0,\ldots,k}$ is absent, the prediction problem described in Observation~\ref{o1} reduces to that of, at time $k$, predicting
$\mathbf{y}_{k}$ based only upon the previous $\mathbf{y}_{0,\ldots,k-1}$, that is
\begin{flalign}
\widehat{\mathbf{y}}_{k} = g_{k} \left( \mathbf{y}_{0,\ldots,k-1} \right),
\end{flalign}
where $\widehat{\mathbf{y}}_{k}$ denotes the prediction of $\mathbf{y}_{k}$.
Accordingly, Theorem~\ref{MIMOFano} reduces to the following Corollary~\ref{MIMOFanoC}.

\begin{corollary} \label{MIMOFanoC}
	For any prediction algorithm $g_{k} \left( \cdot \right)$, it holds that  
	\begin{flalign}
	\left[ \mathbb{E} \left( \left| \mathbf{y}_{k} - \widehat{\mathbf{y}}_{k} \right|^{p} \right) \right]^{\frac{1}{p}}
	\geq \frac{2^{h \left( \mathbf{y}_k | \mathbf{y}_{0,\ldots,k-1} \right)}}{2 \Gamma \left( \frac{p+1}{p} \right) \left( p \mathrm{e} \right)^{\frac{1}{p}}},
	\end{flalign}
	where equality holds if and only if $\mathbf{y}_{k} - \widehat{\mathbf{y}}_{k}$ is with probability density
	\begin{flalign}
	f_{\mathbf{y}_{k} - \widehat{\mathbf{y}}_{k}} \left( y \right)
	= \frac{ \mathrm{e}^{- \left| y \right|^{p} / \left( p \mu^{p} \right)} }{2 \Gamma \left( \frac{p+1}{p} \right) p^{\frac{1}{p}} \mu},
	\end{flalign}
	and $I \left( \mathbf{y}_{k} - \widehat{\mathbf{y}}_{k}; \mathbf{y}_{0,\ldots,k-1} \right) = 0$.  Note that herein
	\begin{flalign}
	\mu 
	= \frac{2^{h \left( \mathbf{y}_k | \mathbf{y}_{0,\ldots,k-1} \right)}}{2 \Gamma \left( \frac{p+1}{p} \right) \left( p \mathrm{e} \right)^{\frac{1}{p}}}.
	\end{flalign}
\end{corollary}

In addition, Proposition~\ref{p2} becomes the following Proposition~\ref{p3}.

\begin{proposition} \label{p3}
	For any $g_{k} \left( \cdot \right)$, it holds that  
	\begin{flalign}
	&I \left( \mathbf{y}_{k} - \widehat{\mathbf{y}}_{k}; \mathbf{y}_{0,\ldots,k-1} \right) \nonumber \\
	&~~~~ = I \left( \mathbf{y}_{k} - \widehat{\mathbf{y}}_{k} ; \mathbf{y}_{0} - \widehat{\mathbf{y}}_{0}, \ldots, \mathbf{y}_{k-1} - \widehat{\mathbf{y}}_{k-1} \right).
	\end{flalign}
\end{proposition}

In other words, the mutual information between the
current innovation $\mathbf{y}_{k} - \widehat{\mathbf{y}}_{k}$ and the previous observations $\mathbf{y}_{0,\ldots,k-1}$ is equal to the mutual information
between the current innovation $\mathbf{y}_{k} - \widehat{\mathbf{y}}_{k}$ and the previous innovations $\mathbf{y}_{0} - \widehat{\mathbf{y}}_{0}, \ldots, \mathbf{y}_{k-1} - \widehat{\mathbf{y}}_{k-1}$.
Accordingly, the condition
\begin{flalign}
I \left( \mathbf{y}_{k} - \widehat{\mathbf{y}}_{k}; \mathbf{y}_{0,\ldots,k-1} \right) = 0
\end{flalign}
is
equivalent to 
\begin{flalign}
I \left( \mathbf{y}_{k} - \widehat{\mathbf{y}}_{k} ; \mathbf{y}_{0} - \widehat{\mathbf{y}}_{0}, \ldots, \mathbf{y}_{k-1} - \widehat{\mathbf{y}}_{k-1} \right) = 0,
\end{flalign}
which in turn means that the current innovation
contains no information of the previous innovations. As such, Corollary~\ref{MIMOasymp} reduces to the following Corollary~\ref{MIMOasymp2}.

\begin{corollary} \label{MIMOasymp2}
	For any prediction algorithm $g_{k} \left( \cdot \right)$, it holds that  
	\begin{flalign}
	\liminf_{k\to \infty} \left[ \mathbb{E} \left( \left| \mathbf{y}_{k} - \widehat{\mathbf{y}}_{k} \right|^{p} \right) \right]^{\frac{1}{p}}
	\geq \liminf_{k\to \infty} \frac{2^{h \left( \mathbf{y}_k | \mathbf{y}_{0,\ldots,k-1} \right)}}{2 \Gamma \left( \frac{p+1}{p} \right) \left( p \mathrm{e} \right)^{\frac{1}{p}}},
	\end{flalign}
	where equality holds if $\left\{ \mathbf{y}_{k} - \widehat{\mathbf{y}}_{k} \right\}$ is asymptotically independent over time, i.e.,
	\begin{flalign}
	\lim_{k \to \infty} I \left( \mathbf{y}_{k} - \widehat{\mathbf{y}}_{k} ; \mathbf{y}_{0} - \widehat{\mathbf{y}}_{0}, \ldots, \mathbf{y}_{k-1} - \widehat{\mathbf{y}}_{k-1} \right) = 0,
	\end{flalign} 
	and with probability density
	\begin{flalign} \label{asydistribution2}
	\lim_{k \to \infty} f_{\mathbf{y}_{k} - \widehat{\mathbf{y}}_{k}} \left( y \right)
	= \frac{ \mathrm{e}^{- \left| y \right|^{p} / \left( p \mu^{p} \right)} }{2 \Gamma \left( \frac{p+1}{p} \right) p^{\frac{1}{p}} \mu}.
	\end{flalign}
	 Note that herein
	\begin{flalign}
	\mu 
	= \liminf_{k\to \infty} \frac{2^{h \left( \mathbf{y}_k | \mathbf{y}_{0,\ldots,k-1} \right)}}{2 \Gamma \left( \frac{p+1}{p} \right) \left( p \mathrm{e} \right)^{\frac{1}{p}}}.
	\end{flalign}
\end{corollary}

When $\left\{ \mathbf{y}_{k} \right\}$ forms an asymptotically stationary process, we arrive at the following result.

\begin{corollary} \label{uniform2}
	Consider an asymptotically stationary stochastic process $\left\{ \mathbf{y}_{k} \right\}, \mathbf{y}_{k} \in \mathbb{R}$. 
	Then, for any prediction algorithm $g_{k} \left( \cdot \right)$, it holds that 
	\begin{flalign} 
	\liminf_{k\to \infty} \left[ \mathbb{E} \left( \left| \mathbf{y}_{k} - \widehat{\mathbf{y}}_{k} \right|^{p} \right) \right]^{\frac{1}{p}}
	\geq \frac{2^{h_{\infty} \left( \mathbf{y} \right)}}{2 \Gamma \left( \frac{p+1}{p} \right) \left( p \mathrm{e} \right)^{\frac{1}{p}}},
	\end{flalign}
	where $h_{\infty} \left( \mathbf{y} \right)$ denotes the entropy rate of $\left\{ \mathbf{y}_{k} \right\}$. Herein, equality holds if $\left\{ \mathbf{y}_{k} - \widehat{\mathbf{y}}_{k} \right\}$ is asymptotically independent over time and with probability density \eqref{asydistribution2}.
\end{corollary}

Corollary~\ref{uniform2} follows directly from Corollary~\ref{MIMOasymp} by noting that for asymptotically stationary processes $\left\{ \mathbf{y}_{k} \right\}$, we have \cite{Cov:06} 
\begin{flalign} 
\liminf_{k\to \infty} h \left( \mathbf{y}_k | \mathbf{y}_{0,\ldots,k-1} \right) = \lim_{k\to \infty} h \left( \mathbf{y}_k | \mathbf{y}_{0,\ldots,k-1} \right)  = h_{\infty} \left( \mathbf{y} \right). \nonumber
\end{flalign}

As a matter of fact, in Corollary~\ref{uniform2}, if $\left\{ \mathbf{y}_{k} - \widehat{\mathbf{y}}_{k} \right\}$ is asymptotically independent over time and with probability density \eqref{asydistribution2}, then, noting also that $\left\{ \mathbf{y}_{k} \right\}$ is asymptotically stationary, it holds that
\begin{flalign} \label{equality2}
\lim_{k\to \infty} \left[ \mathbb{E} \left( \left| \mathbf{y}_{k} - \widehat{\mathbf{y}}_{k} \right|^{p} \right) \right]^{\frac{1}{p}}
= \frac{2^{h_{\infty} \left( \mathbf{y} \right)}}{2 \Gamma \left( \frac{p+1}{p} \right) \left( p \mathrm{e} \right)^{\frac{1}{p}}}.
\end{flalign}
In addition, we can show that \eqref{equality2} holds if and only if $\left\{ \mathbf{y}_{k} - \widehat{\mathbf{y}}_{k} \right\}$ is asymptotically independent over time and with probability density \eqref{asydistribution2}; in other words, the necessary and sufficient condition for achieving the prediction bounds asymptotically is that the innovation is asymptotically independent over time and with probability density \eqref{asydistribution2}.

\subsection{Special Cases}


In this subsection, we first examine in further details the cases of when $p$ is assigned with particular values. 

\begin{corollary} We now consider the special cases of Theorem~\ref{MIMOFano} for when $p=1$, $p=2$, and $p=\infty$, respectively.
	\begin{itemize}
		\item When $p=1$, it holds for any prediction algorithm $g_{k} \left( \cdot \right)$ that
		\begin{flalign} 
		\mathbb{E} \left| \mathbf{y}_{k} - \widehat{\mathbf{y}}_{k} \right|
		\geq \frac{2^{h \left( \mathbf{y}_k | \mathbf{y}_{0,\ldots,k-1}, \mathbf{x}_{0,\ldots,k} \right)}}{2\mathrm{e}},
		\end{flalign}
		where equality holds if and only if $\mathbf{y}_{k} - \widehat{\mathbf{y}}_{k}$ is with probability density 
		\begin{flalign}
		f_{\mathbf{y}_{k} - \widehat{\mathbf{y}}_{k}} \left( y \right)
		= \frac{ \mathrm{e}^{- \left| y \right| / \mu } }{2 \mu},
		\end{flalign}
		that is to say, if and only if $\mathbf{y}_{k} - \widehat{\mathbf{y}}_{k}$ is Laplace,
		and $I \left( \mathbf{y}_{k} - \widehat{\mathbf{y}}_{k}; \mathbf{y}_{0,\ldots,k-1}, \mathbf{x}_{0,\ldots,k} \right) = 0$. Note that herein \begin{flalign}
		\mu 
		= \frac{2^{h \left( \mathbf{y}_k | \mathbf{y}_{0,\ldots,k-1}, \mathbf{x}_{0,\ldots,k} \right)}}{2  \mathrm{e} }.
		\end{flalign}
		\item When $p=2$, it holds for any prediction algorithm $g_{k} \left( \cdot \right)$ that
		\begin{flalign} \label{variance}
		\left\{ \mathbb{E} \left[ \left( \mathbf{y}_{k} - \widehat{\mathbf{y}}_{k} \right)^{2} \right]  \right\}^{\frac{1}{2}}
		\geq \frac{2^{h \left( \mathbf{y}_k | \mathbf{y}_{0,\ldots,k-1}, \mathbf{x}_{0,\ldots,k} \right)}}{\left( 2 \pi \mathrm{e} \right)^{\frac{1}{2}}},
		\end{flalign}
		where equality holds if and only if $\mathbf{y}_{k} - \widehat{\mathbf{y}}_{k}$ is with probability density
		\begin{flalign}
		f_{\mathbf{y}_{k} - \widehat{\mathbf{y}}_{k}} \left( y \right)
		= \frac{ \mathrm{e}^{- y^{2} / \left( 2 \mu^{2} \right)} }{ \left(2 \pi \mu^2 \right)^{\frac{1}{2}} },
		\end{flalign}
		that is to say, if and only if $\mathbf{y}_{k} - \widehat{\mathbf{y}}_{k}$ is Gaussian,
		and $I \left( \mathbf{y}_{k} - \widehat{\mathbf{y}}_{k}; \mathbf{y}_{0,\ldots,k-1}, \mathbf{x}_{0,\ldots,k} \right) = 0$. Note that herein
		\begin{flalign}
		\mu = \frac{2^{h \left( \mathbf{y}_k | \mathbf{y}_{0,\ldots,k-1}, \mathbf{x}_{0,\ldots,k} \right)}}{\left( 2 \pi \mathrm{e} \right)^{\frac{1}{2}}}.
		\end{flalign}
		\item When $p=\infty$, it holds for any prediction algorithm $g_{k} \left( \cdot \right)$ that
		\begin{flalign} \label{MD}
		\esssup_{ f_{\mathbf{y}_{k} - \widehat{\mathbf{y}}_{k}} \left( y \right) > 0} \left| \mathbf{y}_{k} - \widehat{\mathbf{y}}_{k} \right|
		\geq \frac{2^{h \left( \mathbf{y}_k | \mathbf{y}_{0,\ldots,k-1}, \mathbf{x}_{0,\ldots,k} \right)}}{2},
		\end{flalign}
		where equality holds if and only if $\mathbf{y}_{k} - \widehat{\mathbf{y}}_{k}$ is with probability density
		\begin{flalign}
		f_{\mathbf{y}_{k} - \widehat{\mathbf{y}}_{k}} \left( y \right)
		= \left\{ \begin{array}{cc}
		\frac{1}{2 \mu}, & \left| y \right| \leq \mu,\\
		0, & \left| y \right| > \mu,
		\end{array} \right. 
		\end{flalign}
		that is to say, if and only if $\mathbf{y}_{k} - \widehat{\mathbf{y}}_{k}$ is uniform, and $I \left( \mathbf{y}_{k} - \widehat{\mathbf{y}}_{k}; \mathbf{y}_{0,\ldots,k-1}, \mathbf{x}_{0,\ldots,k} \right) = 0$. Note that herein
		\begin{flalign}
		\mu = \frac{2^{h \left( \mathbf{y}_k | \mathbf{y}_{0,\ldots,k-1}, \mathbf{x}_{0,\ldots,k} \right)}}{2}.
		\end{flalign}
	\end{itemize} 
\end{corollary}

It is clear that in order to minimize different $\mathcal{L}_p$ norms of the prediction error, its distributions should be steered to different ones. Meanwhile, the condition $I \left( \mathbf{y}_{k} - \widehat{\mathbf{y}}_{k}; \mathbf{y}_{0,\ldots,k-1}, \mathbf{x}_{0,\ldots,k} \right) = 0$ stays the same for all $p \geq 1$.





We next investigate the connections to estimation counterparts of Fano's inequality \cite{Cov:06}. 

\begin{corollary} \label{Fano}
	Consider a random variable $\mathbf{x} \in \mathbb{R}$ with side information $\mathbf{y} \in \mathbb{R}^n$.
	Then, it holds for any estimator $\widehat{\mathbf{x}} = g \left( \mathbf{y} \right)$ that  
	\begin{flalign} \label{Fano1}
	\left[ \mathbb{E} \left( \left| \mathbf{x} - \widehat{\mathbf{x}} \right|^{p} \right) \right]^{\frac{1}{p}}
	\geq \frac{2^{h \left( \mathbf{x} | \mathbf{y} \right)}}{2 \Gamma \left( \frac{p+1}{p} \right) \left( p \mathrm{e} \right)^{\frac{1}{p}}}.
	\end{flalign}
    In addition, if the side information $\mathbf{y}$ is absent, it follows that
	\begin{flalign} \label{Fano2}
	\left[ \mathbb{E} \left( \left| \mathbf{x} - \widehat{\mathbf{x}} \right|^{p} \right) \right]^{\frac{1}{p}}
	\geq \frac{2^{h \left( \mathbf{x} \right)}}{2 \Gamma \left( \frac{p+1}{p} \right) \left( p \mathrm{e} \right)^{\frac{1}{p}}}.
	\end{flalign}
\end{corollary}

Note that Corollary~\ref{Fano} can be proved by replacing $\mathbf{y}_k$ and $\mathbf{y}_{0,\ldots,k-1}, \mathbf{x}_{0,\ldots,k}$ by $\mathbf{x}$ and $\mathbf{y}$, respectively, in Theorem~\ref{MIMOFano}.

It can then be verified that when $p=2$, \eqref{Fano1} and \eqref{Fano2} reduce to the so-called estimation counterparts to Fano's inequality \cite{Cov:06}:
\begin{flalign}
\left[ \mathbb{E} \left( \left| \mathbf{x} - \widehat{\mathbf{x}} \right|^{2} \right) \right]^{\frac{1}{2}}
\geq \frac{2^{h \left( \mathbf{x} | \mathbf{y} \right)}}{\left( 2 \pi \mathrm{e} \right)^{\frac{1}{2}}};~\left[ \mathbb{E} \left( \left| \mathbf{x} - \widehat{\mathbf{x}} \right|^{2} \right) \right]^{\frac{1}{2}}
\geq \frac{2^{h \left( \mathbf{x} \right)}}{\left( 2 \pi \mathrm{e} \right)^{\frac{1}{2}}}.
\end{flalign}
Meanwhile, for $p=1$ and $p=\infty$, \eqref{Fano1} and \eqref{Fano2} reduce respectively to
\begin{flalign} 
\mathbb{E} \left| \mathbf{x} - \widehat{\mathbf{x}} \right|
\geq \frac{2^{h \left( \mathbf{x} | \mathbf{y} \right)}}{2\mathrm{e}};~\mathbb{E} \left| \mathbf{x} - \widehat{\mathbf{x}} \right|
\geq \frac{2^{h \left( \mathbf{x} \right)}}{2\mathrm{e}},
\end{flalign}
and
\begin{flalign} 
\esssup_{ f_{\mathbf{x} - \widehat{\mathbf{x}}} \left( x \right) > 0} \left| \mathbf{x} - \widehat{\mathbf{x}} \right|
\geq \frac{2^{h \left( \mathbf{x} | \mathbf{y} \right)}}{2};~\esssup_{ f_{\mathbf{x} - \widehat{\mathbf{x}}} \left( x \right) > 0} \left| \mathbf{x} - \widehat{\mathbf{x}} \right|
\geq \frac{2^{h \left( \mathbf{x} \right)}}{2}.
\end{flalign}
In this sense, \eqref{Fano1} and \eqref{Fano2} may be viewed as generalizations of the estimation counterparts to Fano's inequality. On the other hand, \eqref{Fano1} also reinforces the conclusions in \cite{jiao2015justification}.

\subsection{Relation to the Kolomogorov--Szeg\"o Formula}

As a matter of fact, formulae that are more specific than that of Corollary~\ref{uniform2} could be derived when it comes to predicting asymptotically stationary processes.

\begin{corollary} \label{power1}
	Consider an asymptotically stationary stochastic process $\left\{ \mathbf{y}_{k} \right\}, \mathbf{y}_{k} \in \mathbb{R}$ with asymptotic power spectrum $ S_{\mathbf{y}} \left( \omega \right)$, which is defined as \cite{Pap:02}
	\begin{flalign}
	&S_{\mathbf{y}}\left( \omega\right)
	=\sum_{k=-\infty}^{\infty} R_{\mathbf{y}}\left( k\right) \mathrm{e}^{-\mathrm{j}\omega k}, \nonumber 
	\end{flalign}
	and herein
	$R_{\mathbf{y}}\left( k\right) =\lim_{i\to \infty} \mathrm{E}\left[ \mathbf{y}_i \mathbf{y}_{i+k} \right]$ denotes the asymptotic correlation matrix.
	Then, for any prediction algorithm $g_{k} \left( \cdot \right)$, it holds that 
	\begin{flalign} \label{spectrum}
	&\liminf_{k\to \infty} \left[ \mathbb{E} \left( \left| \mathbf{y}_{k} - \widehat{\mathbf{y}}_{k} \right|^{p} \right) \right]^{\frac{1}{p}} \nonumber \\
	&~~~~ \geq \frac{\left[ 2^{- J_{\infty} \left( \mathbf{y} \right)} \right] 2^{\frac{1}{2 \mathrm{\pi}}\int_{-\mathrm{\pi}}^{\mathrm{\pi}}{\log \sqrt{2 \pi \mathrm{e} S_{ \mathbf{y} } \left( \omega \right) } \mathrm{d}\omega }}}{2 \Gamma \left( \frac{p+1}{p} \right) \left( p \mathrm{e} \right)^{\frac{1}{p}}},
	\end{flalign}
	where $J_{\infty} \left( \mathbf{y} \right)$ denotes the negentropy rate \cite{fang2017towards} of $\left\{ \mathbf{y}_{k} \right\}$, $J_{\infty} \left( \mathbf{y} \right) \geq 0$, and $J_{\infty} \left( \mathbf{y} \right) = 0$ if and only if $\left\{ \mathbf{y}_{k} \right\}$ is Gaussian.
	Herein, equality holds if $\left\{ \mathbf{y}_{k} - \widehat{\mathbf{y}}_{k} \right\}$ is asymptotically independent over time and with probability density \eqref{asydistribution2}.
\end{corollary}

\begin{IEEEproof}
Note first that for an asymptotically stationary stochastic process $\left\{ \mathbf{y}_{k} \right\}$ with asymptotic power spectrum $ S_{\mathbf{y}} \left( \omega \right)$, we have \cite{fang2017towards}
\begin{flalign} 
h_{\infty} \left( \mathbf{y} \right) 
= \frac{1}{2 \mathrm{\pi}} \int_{-\mathrm{\pi}}^{\mathrm{\pi}} \log \sqrt{2 \pi \mathrm{e} S_{ \mathbf{y} } \left( \omega \right) } \mathrm{d}\omega - J_{\infty} \left( \mathbf{y} \right).  \nonumber
\end{flalign}
Consequently,
\begin{flalign}
2^{h_{\infty} \left( \mathbf{y} \right)} 
= \left[ 2^{- J_{\infty} \left( \mathbf{y} \right)} \right] 2^{\frac{1}{2 \mathrm{\pi}}\int_{-\mathrm{\pi}}^{\mathrm{\pi}}{\log \sqrt{2 \pi \mathrm{e} S_{ \mathbf{y} } \left( \omega \right) } \mathrm{d}\omega }}. \nonumber
\end{flalign}
This completes the proof.
\end{IEEEproof}

Herein, negentropy rate is a measure of non-Gaussianity for asymptotically stationary sequences, which grows larger as the sequence to be predicted becomes less Gaussian; see \cite{fang2017towards} for more details of its properties. Accordingly, the bounds in \eqref{spectrum} will decrease as $\left\{ \mathbf{y}_{k} \right\}$ becomes less Gaussian, and vice versa. In the limit when  $\left\{ \mathbf{y}_{k} \right\}$ is Gaussian, \eqref{spectrum} reduces to
\begin{flalign} \label{generalizedKS}
\liminf_{k\to \infty} \left[ \mathbb{E} \left( \left| \mathbf{y}_{k} - \widehat{\mathbf{y}}_{k} \right|^{p} \right) \right]^{\frac{1}{p}}
\geq \frac{ 2^{\frac{1}{2 \mathrm{\pi}}\int_{-\mathrm{\pi}}^{\mathrm{\pi}}{\log \sqrt{2 \pi \mathrm{e} S_{ \mathbf{y} } \left( \omega \right) } \mathrm{d}\omega }}}{2 \Gamma \left( \frac{p+1}{p} \right) \left( p \mathrm{e} \right)^{\frac{1}{p}}}.
\end{flalign}
%

In addition, when $p=2$, \eqref{generalizedKS} further reduces to
\begin{flalign}
\liminf_{k\to \infty} \left\{ \mathbb{E} \left[ \left( \mathbf{y}_{k} - \widehat{\mathbf{y}}_{k} \right)^{2} \right]  \right\}^{\frac{1}{2}}
&\geq \frac{ 2^{\frac{1}{2 \mathrm{\pi}}\int_{-\mathrm{\pi}}^{\mathrm{\pi}}{\log \sqrt{2 \pi \mathrm{e} S_{ \mathbf{y} } \left( \omega \right) } \mathrm{d}\omega }}}{\left( 2 \pi \mathrm{e} \right)^{\frac{1}{2}}}
\nonumber \\
&= 2^{\frac{1}{2 \mathrm{\pi}}\int_{-\mathrm{\pi}}^{\mathrm{\pi}}{\log \sqrt{S_{ \mathbf{y} } \left( \omega \right) } \mathrm{d}\omega }}.
\end{flalign}
Meanwhile, we can show that
\begin{flalign}
\lim_{k\to \infty} \left\{ \mathbb{E} \left[ \left( \mathbf{y}_{k} - \widehat{\mathbf{y}}_{k} \right)^{2} \right]  \right\}^{\frac{1}{2}}
&= 2^{\frac{1}{2 \mathrm{\pi}}\int_{-\mathrm{\pi}}^{\mathrm{\pi}}{\log \sqrt{S_{ \mathbf{y} } \left( \omega \right) } \mathrm{d}\omega }},
\end{flalign}
if and only if  $\left\{ \mathbf{y}_{k} - \widehat{\mathbf{y}}_{k} \right\}$ is asymptotically white Gaussian, which coincides with the Kolmogorov--Szeg\"o formula \cite{Pap:02, vaidyanathan2007theory, lindquist2015linear}. In this sense, \eqref{generalizedKS} as well as \eqref{spectrum} may be viewed as generalizations of the Kolmogorov--Szeg\"o formula.

\section{Fundamental Limits of Generalization}

In this section, we examine the fundamental limits of generalization.
In a broad sense, generalization \cite{hastie2009elements, goodfellow2016deep, wolpert2018mathematics} in learning problems (to be more specific, fitting problems) can be viewed as a prediction problem with side information, by noting that the subscript $i$ in Theorem~\ref{MIMOFano} do not necessarily denote time instants, but may more generally denote the indices of the data points. 


More specifically, consider the training data as input/output pairs $\left( \mathbf{x}_{i}, \mathbf{y}_{i} \right)$, $i = 1, \ldots, k$, where $\mathbf{x}_{i} \in \mathbb{R}^n$ denotes the input while $\mathbf{y}_{i} \in \mathbb{R}$ denotes the output. In addition, let the test input/output data pair be denoted as $\left( \mathbf{x}_{\text{test}}, \mathbf{y}_{\text{test}} \right)$. Suppose that the trained learning algorithm (based upon all the training input/output pairs $\left( \mathbf{x}_{i}, \mathbf{y}_{i} \right)$, $i = 1, \ldots, k$), as a mapping from input to output, is denoted as $g \left( \cdot \right)$. Subsequently, $g \left( \cdot \right)$ will be utilized to give a ``prediction" of $\mathbf{y}_{\text{test}}$ with input $\mathbf{x}_{\text{test}}$, and this prediction will be denoted as 
\begin{flalign}
\widehat{\mathbf{y}}_{\text{test}} = g \left( \mathbf{x}_{\text{test}} \right).
\end{flalign}
Observe then that the parameters of $g \left( \cdot \right)$ are trained using $ \left( \mathbf{x}_{i}, \mathbf{y}_{i} \right), i = 1,\ldots,k$, and hence it holds that
\begin{flalign} \label{prediction2}
\widehat{\mathbf{y}}_{\text{test}} = g \left( \mathbf{x}_{\text{test}} \right) = \widehat{g} \left( \mathbf{x}_{\text{test}}, \mathbf{y}_{1,\ldots,k}, \mathbf{x}_{1,\ldots,k} \right),
\end{flalign}
meaning that $\widehat{\mathbf{y}}_{\text{test}}$ is eventually a function, denoted herein as $\widehat{g} \left( \cdot \right)$, of $ \mathbf{x}_{\text{test}}, \mathbf{y}_{1,\ldots,k}, \mathbf{x}_{1,\ldots,k}$. Note that \eqref{prediction2} is essentially a prediction problem with side information, that is, to predict $\mathbf{y}_{\text{test}}$ based on  $\mathbf{y}_{1,\ldots,k}$ with side information $\mathbf{x}_{1,\ldots,k}$ and $\mathbf{x}_{\text{test}}$. 

Then, in light of Theorem~\ref{MIMOFano}, we can similarly derive the following generic lower bound on the $\mathcal{L}_{p}$ norm of the generalization error $\mathbf{y}_{\text{test}} - \widehat{\mathbf{y}}_{\text{test}}$, which is valid for all possible learning algorithms $g \left( \cdot \right)$, as any deterministic or randomized functions/mappings.

\begin{theorem} \label{generalization}
	For any learning algorithm $g \left( \cdot \right)$, it holds that  
	\begin{flalign}
	\left[ \mathbb{E} \left( \left| \mathbf{y}_{\text{test}} - \widehat{\mathbf{y}}_{\text{test}} \right|^{p} \right) \right]^{\frac{1}{p}}
	\geq \frac{2^{h \left( \mathbf{y}_{\text{test}} | \mathbf{x}_{\text{test}}, \mathbf{y}_{1,\ldots,k}, \mathbf{x}_{1,\ldots,k} \right)}}{2 \Gamma \left( \frac{p+1}{p} \right) \left( p \mathrm{e} \right)^{\frac{1}{p}}},
	\end{flalign}
	where equality holds if and only if $\mathbf{y}_{\text{test}} - \widehat{\mathbf{y}}_{\text{test}}$ is with probability density
	\begin{flalign}
	f_{\mathbf{y}_{\text{test}} - \widehat{\mathbf{y}}_{\text{test}}} \left( y \right)
	= \frac{ \mathrm{e}^{- \left| y \right|^{p} / \left( p \mu^{p} \right)} }{2 \Gamma \left( \frac{p+1}{p} \right) p^{\frac{1}{p}} \mu},
	\end{flalign}
	and $I \left( \mathbf{y}_{\text{test}} - \widehat{\mathbf{y}}_{\text{test}}; \mathbf{x}_{\text{test}}, \mathbf{y}_{1,\ldots,k}, \mathbf{x}_{1,\ldots,k} \right) = 0$. Note that herein
	\begin{flalign} 
	\mu 
	= \frac{2^{h \left( \mathbf{y}_{\text{test}} | \mathbf{x}_{\text{test}}, \mathbf{y}_{1,\ldots,k}, \mathbf{x}_{1,\ldots,k} \right)}}{2 \Gamma \left( \frac{p+1}{p} \right) \left( p \mathrm{e} \right)^{\frac{1}{p}}}.
	\end{flalign}
\end{theorem}

Theorem~\ref{generalization} implicates that the generalization error is fundamentally lower bounded by the conditional entropy of $ \mathbf{y}_{\text{test}}$ given $ \mathbf{x}_{\text{test}}$ as well as all the training data $ \mathbf{y}_{1,\ldots,k}, \mathbf{x}_{1,\ldots,k}$.


As a matter of fact, Theorem~\ref{generalization} corresponds to the case of supervised learning \cite{hastie2009elements, goodfellow2016deep}. We next consider the cases of semi-supervised learning \cite{semisupervised} as well as unsupervised learning \cite{hinton1999unsupervised} based upon Theorem~\ref{generalization}.

\begin{corollary} The generalization errors in semi-supervised learning and unsupervised learning are lower bounded as follows.
	\begin{itemize}
		\item For semi-supervised learning, i.e., when $\mathbf{y}_{i}$ are missing for, say, $i=i_1, \ldots, i_n$, it holds for any learning algorithm $g \left( \cdot \right)$ that
		\begin{flalign} \label{semi}
		&\left[ \mathbb{E} \left( \left| \mathbf{y}_{\text{test}} - \widehat{\mathbf{y}}_{\text{test}} \right|^{p} \right) \right]^{\frac{1}{p}} \nonumber \\
		&~~~~ \geq \frac{2^{h \left( \mathbf{y}_{\text{test}} | \mathbf{y}_{1,\ldots,i_1-1, i_1+1, \ldots,i_n-1, i_n+1,\ldots,k}, \mathbf{x}_{1,\ldots,k} \right)}}{2 \Gamma \left( \frac{p+1}{p} \right) \left( p \mathrm{e} \right)^{\frac{1}{p}}}.
		\end{flalign}
		\item For unsupervised learning, i.e., when $\mathbf{y}_{1,\ldots,k}$ are absent, it holds for any learning algorithm $g \left( \cdot \right)$ that
		\begin{flalign} \label{un}
		\left[ \mathbb{E} \left( \left| \mathbf{y}_{\text{test}} - \widehat{\mathbf{y}}_{\text{test}} \right|^{p} \right) \right]^{\frac{1}{p}}
		\geq \frac{2^{h \left( \mathbf{y}_{\text{test}} | \mathbf{x}_{1,\ldots,k} \right)}}{2 \Gamma \left( \frac{p+1}{p} \right) \left( p \mathrm{e} \right)^{\frac{1}{p}}}.
		\end{flalign}
	\end{itemize}	
\end{corollary}

Note that the necessary and sufficient conditions for achieving equalities in \eqref{semi} and \eqref{un} can be analyzed similarly as in Theorem~\ref{generalization}.
Note also that it may be verified that
\begin{flalign} 
&h \left( \mathbf{y}_{\text{test}} | \mathbf{y}_{1,\ldots,k}, \mathbf{x}_{1,\ldots,k} \right)\nonumber \\
&~~~~ \leq h \left( \mathbf{y}_{\text{test}} | \mathbf{y}_{1,\ldots,i_1-1, i_1+1, \ldots,i_n-1, i_n+1,\ldots,k}, \mathbf{x}_{1,\ldots,k} \right) \nonumber \\
&~~~~ \leq h \left( \mathbf{y}_{\text{test}} | \mathbf{x}_{1,\ldots,k} \right),
\end{flalign}
meaning that in general the generalization error of supervised learning is less than or equal to that of semi-supervised learning, which is turn is less than or equal to that of unsupervised learning.

It is worth mentioning that for the generalization error bounds (as well as the previous prediction bounds) obtained in this paper, the classes of learning algorithms (prediction algorithms) that can be applied are not restricted in general, allowing them to be any deterministic or randomized functions/mappings. This means that the bounds are valid for any learning algorithms (prediction algorithms) in practical use, from classical regression methods to deep learning. Note also that no specific restrictions have been imposed on the distributions of the data points either; the data points are not necessarily i.i.d., for instance.

Meanwhile, the fundamental lower bounds may feature baselines for performance assessment and evaluation for various learning algorithms (prediction algorithms), by providing ``best-case" bounds that are to be compared with the true performances. Such baselines may function as fundamental benchmarks that separate what is possible and what is impossible, and can thus be applied to indicate how much room is left for performance improvement, or to avoid infeasible performance specifications in the first place, saving time to be spent on unnecessary parameter tuning work that is destined to be futile.
In addition, those ``best-case" bounds may also inspire learning (prediction) algorithm design. This is enabled by further examining the necessary and/or sufficient conditions for achieving the performance bounds, so as to turn them into optimality conditions and even objective functions for the optimization problems formulated accordingly. Such problems, however, already go beyond the scope of this paper and are potential future research topics.

\section{Fundamental Limits of Recursion}

In this section, we analyze the fundamental limits of recursive algorithms \cite{kushner2003stochastic}, by viewing them as generalized prediction problems.
More specifically, consider a recursive algorithm given by 
\begin{flalign} \label{recursivesystem}
\mathbf{y}_{k+1} = \mathbf{y}_{k} + g_{k} \left( \mathbf{y}_{0,\ldots,k} \right) + \mathbf{n}_{k},
\end{flalign} 
where $\mathbf{y}_{k} \in \mathbb{R}$ denotes the recursive state, $\mathbf{n}_{k} \in \mathbb{R}$ denotes the noise, and $g_{k} \left( \mathbf{y}_{0,\ldots,k} \right) \in \mathbb{R}$ denotes the recursion function.
The following generic bound on the $\mathcal{L}_{p}$ norm of the recursive difference $\mathbf{y}_{k+1} - \mathbf{y}_{k}$ can be obtained.

\begin{theorem} \label{recursivetheorem}
	For any recursion function $g_{k} \left( \cdot \right)$, it holds that
	\begin{flalign} \label{recursion}
	\left[ \mathbb{E} \left( \left| \mathbf{y}_{k+1} - \mathbf{y}_{k} \right|^{p} \right) \right]^{\frac{1}{p}}
	\geq \frac{2^{h \left( \mathbf{n}_k | \mathbf{n}_{0,\ldots,k-1}, \mathbf{y}_{0} \right)}}{2 \Gamma \left( \frac{p+1}{p} \right) \left( p \mathrm{e} \right)^{\frac{1}{p}}},
	\end{flalign}
	where equality holds if and only if $\mathbf{y}_{k+1} - \mathbf{y}_{k}$ is with probability density 
	\begin{flalign} \label{recursivedistribution}
	f_{\mathbf{y}_{k+1} - \mathbf{y}_{k}} \left( x \right)
	= \frac{ \mathrm{e}^{- \left| x \right|^{p} / \left( p \mu^{p} \right)} }{2 \Gamma \left( \frac{p+1}{p} \right) p^{\frac{1}{p}} \mu},
	\end{flalign}
	and $I \left( \mathbf{y}_{k+1} - \mathbf{y}_{k}; \mathbf{n}_{0,\ldots,k-1}, \mathbf{y}_{0} \right) = 0$. Note that herein
	\begin{flalign} 
	\mu 
	= \frac{2^{h \left( \mathbf{n}_k | \mathbf{n}_{0,\ldots,k-1}, \mathbf{y}_{0} \right)}}{2 \Gamma \left( \frac{p+1}{p} \right) \left( p \mathrm{e} \right)^{\frac{1}{p}}}.
	\end{flalign}
\end{theorem}

Before we prove Theorem~\ref{recursivetheorem}, we first prove the following proposition.

\begin{proposition} \label{function1}
	For the recursive algorithm given in \eqref{recursivesystem}, it holds that $\mathbf{y}_{k}$ is eventually a function of $\mathbf{n}_{0,\ldots,k-1}$ and $\mathbf{y}_{0}$, i.e., 
	\begin{flalign} \label{function}
	\mathbf{y}_{k} = l_{k} \left( \mathbf{n}_{0,\ldots,k-1}, \mathbf{y}_{0} \right).
	\end{flalign} 
\end{proposition}

\begin{IEEEproof}To begin with, it is clear that when $k=0$, \eqref{recursivesystem} reduces to  
\begin{flalign}
\mathbf{y}_{1} = \mathbf{y}_{0} + g_{0} \left( \mathbf{y}_{0} \right) + \mathbf{n}_{0}, \nonumber
\end{flalign}
and thus it holds that 
\begin{flalign}
\mathbf{y}_{1} = l_{0} \left( \mathbf{n}_{0}, \mathbf{y}_{0} \right), \nonumber
\end{flalign}
that is, \eqref{function} holds for $k=0$. Next, when $k=1$, 
\eqref{recursivesystem} is given by 
\begin{flalign}
\mathbf{y}_{2} = \mathbf{y}_{1} + g_{1} \left( \mathbf{y}_{0}, \mathbf{y}_{1} \right) + \mathbf{n}_{1}. \nonumber
\end{flalign}
As such, since $\mathbf{y}_{1}$ is a function of $\mathbf{n}_{0}$ and $\mathbf{y}_{0}$, we have 
\begin{flalign}
\mathbf{y}_{2} = l_{0} \left( \mathbf{n}_{0}, \mathbf{y}_{0} \right) + g_{1} \left( \mathbf{y}_{0}, l_{0} \left( \mathbf{n}_{0}, \mathbf{y}_{0} \right) \right) + \mathbf{n}_{1}. \nonumber
\end{flalign} 
In other words, $\mathbf{y}_{2}$ is a function of $\mathbf{n}_{0,1}$ and $\mathbf{y}_{0}$, and thus \eqref{function} holds for $k=1$. We may then repeat this process and show that \eqref{function} holds for any $k \geq 0$.
\end{IEEEproof}

We next prove Theorem~\ref{recursivetheorem} based upon Proposition~\ref{function1}.

\begin{IEEEproof}
It is known from Lemma~\ref{maximum} that 
\begin{flalign}
\left[ \mathbb{E} \left( \left| \mathbf{y}_{k+1} - \mathbf{y}_{k} \right|^{p} \right) \right]^{\frac{1}{p}}
\geq \frac{2^{h \left(  \mathbf{y}_{k+1} - \mathbf{y}_{k} \right)}}{2 \Gamma \left( \frac{p+1}{p} \right) \left( p \mathrm{e} \right)^{\frac{1}{p}}}, \nonumber
\end{flalign} 
where equality holds if and only if $\mathbf{y}_{k+1} - \mathbf{y}_{k}$ is with probability density \eqref{recursivedistribution}.
Meanwhile,
\begin{flalign} 
&h \left( \mathbf{y}_{k+1} - \mathbf{y}_{k} \right) \nonumber \\
& = h \left( \mathbf{y}_{k+1} - \mathbf{y}_{k} |  \mathbf{n}_{0,\ldots,k-1}, \mathbf{y}_{0} \right) + I \left( \mathbf{y}_{k+1} - \mathbf{y}_{k}; \mathbf{n}_{0,\ldots,k-1}, \mathbf{y}_{0} \right) \nonumber \\
& = h \left( g_{k} \left( \mathbf{y}_{0,\ldots,k} \right) + \mathbf{n}_{k} |  \mathbf{n}_{0,\ldots,k-1}, \mathbf{y}_{0} \right) \nonumber \\
&~~~~ + I \left( \mathbf{y}_{k+1} - \mathbf{y}_{k}; \mathbf{n}_{0,\ldots,k-1}, \mathbf{y}_{0} \right). \nonumber
\end{flalign}
Then, due to Proposition~\ref{function1}, $g_{k} \left( \mathbf{y}_{0,\ldots,k} \right)$ is a function of $\mathbf{n}_{0,\ldots,k-1}$ and $\mathbf{y}_{0}$. Hence,
\begin{flalign} 
h \left( g_{k} \left( \mathbf{y}_{0,\ldots,k} \right) + \mathbf{n}_{k} |  \mathbf{n}_{0,\ldots,k-1}, \mathbf{y}_{0} \right) = h \left( \mathbf{n}_{k} |  \mathbf{n}_{0,\ldots,k-1}, \mathbf{y}_{0} \right). \nonumber
\end{flalign}
As a result,
$
2^{ h \left( \mathbf{y}_{k+1} - \mathbf{y}_{k} \right)} 
\geq 2^{h \left( \mathbf{n}_k | \mathbf{n}_{0,\ldots,k-1}, \mathbf{y}_{0} \right)},
$
where equality holds if and only if $I \left( \mathbf{y}_{k+1} - \mathbf{y}_{k}; \mathbf{n}_{0,\ldots,k-1}, \mathbf{y}_{0} \right) = 0$.
Therefore,
\begin{flalign}
\left[ \mathbb{E} \left( \left| \mathbf{y}_{k+1} - \mathbf{y}_{k} \right|^{p} \right) \right]^{\frac{1}{p}}
\geq \frac{2^{h \left( \mathbf{n}_k | \mathbf{n}_{0,\ldots,k-1}, \mathbf{y}_{0} \right)}}{2 \Gamma \left( \frac{p+1}{p} \right) \left( p \mathrm{e} \right)^{\frac{1}{p}}}, \nonumber
\end{flalign}
where equality holds if and only if $\mathbf{y}_{k+1} - \mathbf{y}_{k}$ is with probability density \eqref{recursivedistribution} and $I \left( \mathbf{y}_{k+1} - \mathbf{y}_{k}; \mathbf{n}_{0,\ldots,k-1}, \mathbf{y}_{0} \right) = 0$. This completes the proof.
\end{IEEEproof}

It is clear that in Theorem~\ref{recursivetheorem} the lower bound is determined completely by the conditional entropy of the current noise $\mathbf{n}_{k}$ conditioned on the past noises $\mathbf{n}_{0,\ldots,k-1}$ and the initial state of the recursive algorithm.
In addition, if $\mathbf{y}_{0}$ is chosen deterministically, then \eqref{recursion} reduces to
\begin{flalign} 
\left[ \mathbb{E} \left( \left| \mathbf{y}_{k+1} - \mathbf{y}_{k} \right|^{p} \right) \right]^{\frac{1}{p}}
\geq \frac{2^{h \left( \mathbf{n}_k | \mathbf{n}_{0,\ldots,k-1} \right)}}{2 \Gamma \left( \frac{p+1}{p} \right) \left( p \mathrm{e} \right)^{\frac{1}{p}}}.
\end{flalign}

The next corollary examines the asymptotic case.

\begin{corollary} 
	For any recursion function $g_{k} \left( \cdot \right)$, it holds that
	\begin{flalign} \label{recursive}
	\liminf_{k \to \infty} \left[ \mathbb{E} \left( \left| \mathbf{y}_{k+1} - \mathbf{y}_{k} \right|^{p} \right) \right]^{\frac{1}{p}}
	\geq \liminf_{k \to \infty} \frac{2^{h \left( \mathbf{n}_k | \mathbf{n}_{0,\ldots,k-1}, \mathbf{y}_{0} \right)}}{2 \Gamma \left( \frac{p+1}{p} \right) \left( p \mathrm{e} \right)^{\frac{1}{p}}},
	\end{flalign}
	where equality holds if $\left\{ \mathbf{y}_{k+1} - \mathbf{y}_{k} \right\}$ is asymptotically with probability density 
	\begin{flalign} \label{recursivedistribution2}
	\liminf_{k \to \infty} f_{\mathbf{y}_{k+1} - \mathbf{y}_{k}} \left( x \right)
	= \frac{ \mathrm{e}^{- \left| x \right|^{p} / \left( p \mu^{p} \right)} }{2 \Gamma \left( \frac{p+1}{p} \right) p^{\frac{1}{p}} \mu},
	\end{flalign}
	and $\lim_{k \to \infty} I \left( \mathbf{y}_{k+1} - \mathbf{y}_{k}; \mathbf{n}_{0,\ldots,k-1}, \mathbf{y}_{0} \right) = 0$. Note that herein
	\begin{flalign} 
	\mu 
	= \liminf_{k \to \infty} \frac{2^{h \left( \mathbf{n}_k | \mathbf{n}_{0,\ldots,k-1}, \mathbf{y}_{0} \right)}}{2 \Gamma \left( \frac{p+1}{p} \right) \left( p \mathrm{e} \right)^{\frac{1}{p}}}.
	\end{flalign}
\end{corollary}

We now view the term $I \left( \mathbf{y}_{k+1} - \mathbf{y}_{k}; \mathbf{n}_{0,\ldots,k-1}, \mathbf{y}_{0} \right)$ from the perspective of recursive differences, which are essentially the ``entropic innovations" for recursive algorithms.

\begin{proposition} For any $g_{k} \left( \cdot \right)$, it holds that
	\begin{flalign}
	&I \left( \mathbf{y}_{k+1} - \mathbf{y}_{k}; \mathbf{n}_{0,\ldots,k-1}, \mathbf{y}_{0} \right) \nonumber \\
	&~~~~ = I \left( \mathbf{y}_{k+1} - \mathbf{y}_{k} ; \mathbf{y}_{0}, \mathbf{y}_{1} - \mathbf{y}_{0}, \ldots, \mathbf{y}_{k} - \mathbf{y}_{k-1} \right).
	\end{flalign}
\end{proposition}

\begin{IEEEproof}
Since $\mathbf{y}_{k} = \mathbf{y}_{k-1} + g_{k-1} \left( \mathbf{y}_{0,\ldots,k-1} \right) + \mathbf{n}_{k-1}$, we have 
\begin{flalign} 
&I \left( \mathbf{y}_{k+1} - \mathbf{y}_{k} ; \mathbf{n}_{0,\ldots,k-1}, \mathbf{y}_{0} \right) \nonumber \\
& = I \left( \mathbf{y}_{k+1} - \mathbf{y}_{k} ; \mathbf{n}_{0,\ldots,k-2}, \mathbf{y}_{k} - \mathbf{y}_{k-1} - g_{k-1} \left( \mathbf{y}_{0,\ldots,k-1} \right), \mathbf{y}_{0} \right).\nonumber
\end{flalign}
On the other hand, due to Proposition~\ref{function1}, $g_{k-1} \left( \mathbf{y}_{0,\ldots,k-1} \right)$ is a function of $\mathbf{n}_{0,\ldots,k-2}$ and $\mathbf{y}_{0}$. As such,
\begin{flalign} 
&I \left( \mathbf{y}_{k+1} - \mathbf{y}_{k} ; \mathbf{n}_{0,\ldots,k-2}, \mathbf{y}_{k} - \mathbf{y}_{k-1} - g_{k-1} \left( \mathbf{y}_{0,\ldots,k-1} \right), \mathbf{y}_{0} \right) \nonumber \\
& = I \left( \mathbf{y}_{k+1} - \mathbf{y}_{k} ; \mathbf{n}_{0,\ldots,k-2}, \mathbf{y}_{k} - \mathbf{y}_{k-1}, \mathbf{y}_{0} \right)
.\nonumber
\end{flalign}
We may repeat the previous steps until we eventually arrive at
\begin{flalign} 
&I \left( \mathbf{y}_{k+1} - \mathbf{y}_{k} ; \mathbf{n}_{0,\ldots,k-2}, \mathbf{y}_{k} - \mathbf{y}_{k-1} - g_{k-1} \left( \mathbf{y}_{0,\ldots,k-1} \right), \mathbf{y}_{0} \right) \nonumber \\
& = I \left( \mathbf{y}_{k+1} - \mathbf{y}_{k} ; \mathbf{n}_{0,\ldots,k-2}, \mathbf{y}_{k} - \mathbf{y}_{k-1}, \mathbf{y}_{0} \right) \nonumber \\
& = \cdots = I \left( \mathbf{y}_{k+1} - \mathbf{y}_{k} ;  \mathbf{y}_{1} - \mathbf{y}_{0}, \ldots, \mathbf{y}_{k} - \mathbf{y}_{k-1}, \mathbf{y}_{0} \right)
,\nonumber
\end{flalign}
which completes the proof. \end{IEEEproof}

Accordingly, the condition 
\begin{flalign} 
\lim_{k \to \infty} I \left( \mathbf{y}_{k+1} - \mathbf{y}_{k}; \mathbf{n}_{0,\ldots,k-1}, \mathbf{y}_{0} \right)  = 0 
\end{flalign}
is equivalent to 
\begin{flalign} 
\lim_{k \to \infty} I \left( \mathbf{y}_{k+1} - \mathbf{y}_{k} ; \mathbf{y}_{0}, \mathbf{y}_{1} - \mathbf{y}_{0}, \ldots, \mathbf{y}_{k} - \mathbf{y}_{k-1} \right) = 0,
\end{flalign}
indicating that the recursive difference $\left\{ \mathbf{y}_{k+1} - \mathbf{y}_{k} \right\}$ is asymptotically independent over time.


In fact, we may more generally prove the following result.

\begin{theorem} \label{general}
	Consider a recursive algorithm given by 
	\begin{flalign}
	r_{k+1} \left( \mathbf{y}_{0,\ldots,k+1} \right) = g_{k} \left( \mathbf{y}_{0,\ldots,k} \right) + \mathbf{n}_{k},
	\end{flalign} 
	where $r_{k+1} \left( \mathbf{y}_{0,\ldots,k+1} \right), g_{k} \left( \mathbf{y}_{0,\ldots,k} \right) \in \mathbb{R}$, and $\mathbf{n}_{k} \in \mathbb{R}$ denotes the noise. Then,  
	\begin{flalign} \label{general2}
	\left[ \mathbb{E} \left( \left| r_{k+1} \left( \mathbf{y}_{0,\ldots,k+1} \right) \right|^{p} \right) \right]^{\frac{1}{p}}
	\geq \frac{2^{h \left( \mathbf{n}_k | \mathbf{n}_{0,\ldots,k-1}, \mathbf{y}_{0} \right)}}{2 \Gamma \left( \frac{p+1}{p} \right) \left( p \mathrm{e} \right)^{\frac{1}{p}}},
	\end{flalign}
	where equality holds if and only if $r_{k+1} \left( \mathbf{y}_{0,\ldots,k+1} \right)$ is with probability density 
	\begin{flalign}
	f_{r_{k+1} \left( \mathbf{y}_{0,\ldots,k+1} \right)} \left( x \right)
	= \frac{ \mathrm{e}^{- \left| x \right|^{p} / \left( p \mu^{p} \right)} }{2 \Gamma \left( \frac{p+1}{p} \right) p^{\frac{1}{p}} \mu},
	\end{flalign}
	and $I \left( r_{k+1} \left( \mathbf{y}_{0,\ldots,k+1} \right); \mathbf{n}_{0,\ldots,k-1}, \mathbf{y}_{0} \right) = 0$. Note that herein
	\begin{flalign} 
	\mu 
	= \frac{2^{h \left( \mathbf{n}_k | \mathbf{n}_{0,\ldots,k-1}, \mathbf{y}_{0} \right)}}{2 \Gamma \left( \frac{p+1}{p} \right) \left( p \mathrm{e} \right)^{\frac{1}{p}}}.
	\end{flalign}
\end{theorem}

Note that Theorem~\ref{general} can be proved by following similar procedures to those in the proof of Theorem~\ref{recursivetheorem}, to be more specific, by replacing $\mathbf{y}_{k+1} - \mathbf{y}_{k}$ with more generally $r_{k+1} \left( \mathbf{y}_{0,\ldots,k+1} \right)$ therein.

On the other hand, if we let 
\begin{flalign}
r_{k+1} \left( \mathbf{y}_{0,\ldots,k+1} \right) = \mathbf{y}_{k+1} - \mathbf{y}_{k}
\end{flalign} 
in Theorem~\ref{general}, then it reduces to Theorem~\ref{recursivetheorem}. In addition, we present two other examples as follows.
\begin{itemize}
	\item When 
	\begin{flalign}
	r_{k+1} \left( \mathbf{y}_{0,\ldots,k+1} \right) = \mathbf{y}_{k+1},
	\end{flalign} 
	which corresponds to the recursive algorithm
	\begin{flalign}
	\mathbf{y}_{k+1} = g_{k} \left( \mathbf{y}_{0,\ldots,k} \right) + \mathbf{n}_{k},
	\end{flalign} 
	the lower bound in \eqref{general2} becomes
	\begin{flalign}
	\left[ \mathbb{E} \left( \left| \mathbf{y}_{k+1} \right|^{p} \right) \right]^{\frac{1}{p}}
	\geq \frac{2^{h \left( \mathbf{n}_k | \mathbf{n}_{0,\ldots,k-1}, \mathbf{y}_{0} \right)}}{2 \Gamma \left( \frac{p+1}{p} \right) \left( p \mathrm{e} \right)^{\frac{1}{p}}}.
	\end{flalign}
	\item When
	\begin{flalign}
	r_{k+1} \left( \mathbf{y}_{0,\ldots,k+1} \right) = \mathbf{y}_{k+1} - 2 \mathbf{y}_{k} + \mathbf{y}_{k-1},
	\end{flalign} 
	corresponding to the recursive algorithm
	\begin{flalign}
	\mathbf{y}_{k+1} = 2 \mathbf{y}_{k} - \mathbf{y}_{k-1} + g_{k} \left( \mathbf{y}_{0,\ldots,k} \right) + \mathbf{n}_{k},
	\end{flalign} 
	the lower bound in \eqref{general2} reduces to
	\begin{flalign}
	\left[ \mathbb{E} \left( \left| \mathbf{y}_{k+1} - 2 \mathbf{y}_{k} + \mathbf{y}_{k-1} \right|^{p} \right) \right]^{\frac{1}{p}}
	\geq \frac{2^{h \left( \mathbf{n}_k | \mathbf{n}_{0,\ldots,k-1}, \mathbf{y}_{0} \right)}}{2 \Gamma \left( \frac{p+1}{p} \right) \left( p \mathrm{e} \right)^{\frac{1}{p}}}.
	\end{flalign}
\end{itemize}

We may as well consider other classes of $r_{k+1} \left( \mathbf{y}_{0,\ldots,k+1} \right)$ and analyze similarly.


\section{Conclusion}

In this paper, we have presented the fundamental limits of prediction by an information-theoretic approach. The obtained fundamental $\mathcal{L}_{p}$ bounds on prediction errors are seen to be valid for any prediction algorithms, while the data points can be with arbitrary distributions. We have also examined the implications of the results in generalization as well as recursive algorithms.
Possible future research directions include investigating further the conditions to achieve the fundamental limits to see what they might implicate in algorithm design and/or parameter tuning.


%


\balance
\bibliographystyle{IEEEtran}
\bibliography{references}
\end{document}